\documentclass[letterpaper]{article} 
\usepackage{aaai2026}  
\usepackage{times}  
\usepackage{helvet}  
\usepackage{courier}  
\usepackage[hyphens]{url}  
\usepackage{graphicx} 
\usepackage{natbib}  
\usepackage{caption} 
\usepackage{algorithm}
\usepackage{algorithmic}
\usepackage{newfloat}
\usepackage{listings}
\usepackage{graphicx}
\usepackage{multirow}
\usepackage{float}
\usepackage{mathptmx} 
\usepackage{cite}
\usepackage{amsmath,amsfonts}
\usepackage{graphicx}
\usepackage{epstopdf}
\usepackage{float}
\usepackage{textcomp}
\usepackage{xcolor}
\usepackage{xspace}
\usepackage{url}
\usepackage{multirow}
\usepackage{multicol}
\usepackage{colortbl}
\usepackage{tabularx} 
\usepackage{booktabs} 
\usepackage{subcaption} 

\DeclareCaptionStyle{ruled}{labelfont=normalfont,labelsep=colon,strut=off} 
\floatstyle{ruled}
\newfloat{listing}{tb}{lst}{}
\floatname{listing}{Listing}
\ifodd 0
\newcommand{\rev}[1]{{\color{blue}#1}} 
\newcommand{\newrev}[1]{{\color{red}#1}} 
\else
\newcommand{\rev}[1]{#1}
\newcommand{\newrev}[1]{#1} 
\fi

\title{Branch, or Layer? Zeroth-Order Optimization for Continual Learning of Vision-Language Models}
\author{
    Ziwei Liu\textsuperscript{\rm 1,\rm 3}\equalcontrib,
    Borui Kang\textsuperscript{\rm 2}\equalcontrib,
    Wei Li\textsuperscript{\rm 1},
    Hangjie Yuan\textsuperscript{\rm 4}, 
    Yanbing Yang\textsuperscript{\rm 1},\\
    Wenbin Li\textsuperscript{\rm 2},
    Yifan Zhu\textsuperscript{\rm 5},
    Tao Feng\textsuperscript{\rm 6}\thanks{Corresponding author.},
    Jun Luo\textsuperscript{\rm 3}
}
\affiliations{
    \textsuperscript{\rm 1}College of Computer Science, Sichuan University, China \quad \\
    \textsuperscript{\rm 2}School of Computer Science, Nanjing University, China \quad \\
    \textsuperscript{\rm 3}College of Computing and Data Science, Nanyang Technological University, Singapore \quad \\
    \textsuperscript{\rm 4}DAMO Academy, Alibaba Group, China \quad \\
    \textsuperscript{\rm 5}College of Computer Science, Beijing University of Posts and Telecommunications, China \quad \\
    \textsuperscript{\rm 6}Department of Computer Science and Technology, Tsinghua University, China \\
}

\usepackage{bibentry}

\begin{document}

\maketitle

\begin{abstract}
Vision-Language Continual Learning (VLCL) has attracted significant research attention for its robust capabilities, and the adoption of Parameter-Efficient Fine-Tuning (PEFT) strategies is enabling these models to achieve competitive performance with substantially reduced resource consumption.
However, dominated First-Order (FO) optimization is prone to 
trap models in suboptimal local minima, especially in limited exploration subspace within PEFT. To overcome this challenge, this paper pioneers a systematic exploration of adopting Zeroth-Order (ZO) optimization for PEFT-based VLCL.
We first identify the incompatibility of naive full-ZO adoption in VLCL due to optimization process instability.
We then investigate the application of ZO optimization from a modality branch-wise to a fine-grained layer-wise across various training units to identify an optimal strategy. Besides, a key theoretical insight reveals that vision modality exhibit higher variance than language counterparts in VLCL during the ZO optimization process, and we propose a modality-aware ZO strategy, which adopts gradient sign normalization in ZO and constrains vision modality perturbation to further improve performance. Benefiting from the adoption of ZO optimization, PEFT-based VLCL fulfills better ability to escape local minima during the optimization process, extensive experiments on four benchmarks demonstrate that our method achieves state-of-the-art results.
\end{abstract}

\section{Introduction}

Continual Learning (CL) has witnessed significant advancements in convolutional architectures (e.g., ResNet \cite{erd,res_2, pose} and ViT\cite{vit_1,vit_2,vit_3}). 
Recently, Vision-Language Models-based Continual Learning (VLCL) approaches have attracted growing research attention. 
Particularly, CLIP-based methods \cite{clip_cl2,lwf-vr,zhao2,mod-x} have demonstrated robust continual learning capabilities. However, these methods require full-parameter fine-tuning of CLIP models, which incurs substantial computational overhead. \newrev{To overcome this critical bottleneck, Parameter-Efficient Fine-Tuning (PEFT) strategies \cite{peft, peft1,peft2,dmnsp} have recently emerged as a compelling alternative. These techniques make it possible to achieve competitive CL performance with significantly reduced resource consumption. For instance, the VLCL method MoE4Adapter \cite{MoE4Adapters} leverages a PEFT approach to address this limitation.}

\rev{Existing VLCL methods predominantly employ First-Order (FO) optimization strategy \cite{overview_opti}, which update parameters using precise gradients derived from backpropagation. While valued for their stable directional guidance, this approach becomes a drawback in the context of PEFT. Its deterministic update paths limit exploration during training , and the low-dimensional subspace to which PEFT confines optimization makes these methods susceptible to converging to sharp local optima that overfit the current task \cite{SharpMinima, sam}, leading to a performance drop when faced with new tasks and potentially exacerbating catastrophic forgetting.
To explore solutions with stronger generalization capabilities, Zeroth-Order (ZO) optimization offers a promising alternative \cite{zeroflow}. Unlike traditional FO optimization, this method forgoes precise gradients from backpropagation, instead estimating performance with random perturbations, making it less prone to getting trapped in local minima when exploring a constrained space \cite{mezo,maskzo}, hence potentially applied in PEFT-based VLCL.}

\rev{There is already a vast amount of ZO finetuning researches \cite{mezo,maskzo}, while they lacks of considering its common strategy of fully replacing FO optimization is applicable to VLCL, given the documented optimization disparities between modality branches \cite{vlgap,clap4clip,vldis1,vldis2, vldis3} and the inherent sensitivity of ZO's perturbation-based approach.}
In this paper, we initiate the study of applying ZO optimization into VLCL, and aim to answer the following key question:
\textit{How can ZO be integrated in VLCL settings, and can it boost overall performance?}

To answer the question, this paper adopts a fine-grained perspective, investigating the performance of ZO when applied to different modality branches and trainable layers within PEFT-based VLCL. Specifically, we respectively explore ZO in the vision or language modality branch while retaining FO in the other, to identify their benefits and limitations. Based on the insights gained, we then extend our investigation to a more granular layer-wise and adopt ZO into different training units, including continuous \textit{prefix/suffix} layers and interleaved \textit{odd/even} layers across a modality branch, to obtain a optimal result. Meanwhile, we identify a convergence discrepancy between modalities of VLCL under ZO optimization, thus proposing a \textbf{Mo}dality-aware \textbf{ZO} (\textbf{MoZO}) strategy, adopting gradient sign normalization in ZO and constraining vision modality perturbation to further boost VLCL performance.

\begin{figure*}[t!]
    \centering
\includegraphics[width=.99\linewidth]{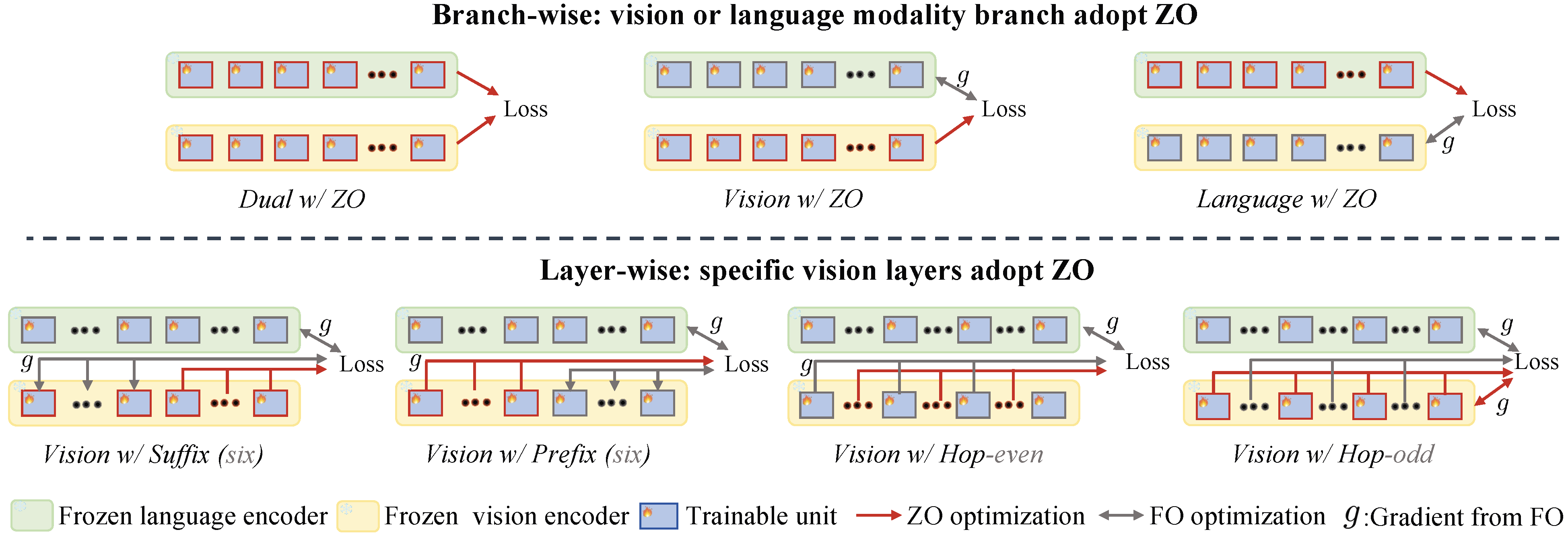}
    \caption{Illustration of our study. The language and vision encoders of \textbf{\texttt{CLIP}} are frozen, only the trainable units attached to each layer is performed to parameters update. To sum up, we systematically explores how ZO optimization operates in VLCL, including branches (\textit{Dual}, \textit{Vision}, or \textit{Language}) and layers (\textit{w/ Hop-\textcolor{gray}{odd}}, \textit{w/ Hop-\textcolor{gray}{even}}, \textit{w/ Prefix (\textcolor{gray}{six}}) and \textit{w/ Suffix (\textcolor{gray}{six}})).}
    \label{fig:arch}
\end{figure*}

In summary, the main contributions of this work are:
\begin{itemize}
    \item We present the optimization challenge in PEFT-based VLCL, where conventional FO method is prone to converging to suboptimal local optima, and explore to leverage the ZO optimization to address the problem.

    \item We adopt the ZO optimization at both branch-wise and fine-grained layer-wise within PEFT-based VLCL, achieving optimal performance through detailed analysis and refined application strategies.

    \item We identify the issue of optimization discrepancy between modalities of PEFT-based VLCL under ZO optimization, and propose a \textbf{MoZO} strategy further improving overall performance.
\end{itemize}

\section{A Preliminary for ZO Optimization in PEFT-based VLCL}

\rev{Traditional optimization methods in continual learning primarily rely on FO gradient descent \cite{fo_1,fo_2}, updating model parameters $\theta$ based on precise gradients computed via backpropagation:
\begin{equation}
    \theta_{t+1} = \theta_t - \eta_t \cdot \nabla_{\theta} \mathcal{L}(\theta_t),
\end{equation}
where $\nabla_{\theta} \mathcal{L}(\theta_t)$ is the first-order gradient. While FO methods offer accurate gradient directions and have been widely, their deterministic update paths tend to limit exploration during training, potentially increasing the risk of convergence to suboptimal local minima and reducing adaptability in dynamic continual learning scenarios.}

\rev{In contrast, ZO optimization estimates gradients through forward passes through purposeful perturbations.} For a parameter subset \(\theta^k_m\), ZO approximates gradients via directional perturbations:  
\begin{equation}
\nabla_{\text{ZO}} \mathcal{L}(\theta^k_m) \approx \frac{\mathcal{L}(\theta^k_m + \varepsilon \Delta) - \mathcal{L}(\theta^k_m)}{\varepsilon} \cdot \Delta,
 \end{equation}
where \(\Delta\) is a random directional vector, and \(\varepsilon\) is a small perturbation scale. \rev{ZO methods introduce gradient stochasticity \cite{derivative}, which may improve the ability to explore non-convex loss landscapes and escape poor local minima. However, the reliance on randomized perturbations can lead to variance in gradient estimates, whose effect may vary across different model architectures or modalities. 
To validate this point, we first establish a baseline and investigate the most straightforward strategy: a complete replacement of the FO optimizer with ZO in PEFT-based VLCL, which means adopt ZO in both vision-language branches and overall trainable units. However, our subsequent experiments demonstrate this strategy leads to severe training instability, evidenced by loss oscillations during the convergence process, and results in a significant performance degradation.
}

\rev{Recent research suggests that partially incorporating exploratory while high-variance ZO estimators into the model architecture improves its global optimization performance \cite{hybridzo1,hybridzo2}.
We translate this advance to the context of PEFT-based VLCL.
As shown in Figure~\ref{fig:arch}, we adopt a empirical exploration to progressively investigate how to best apply the ZO optimization in VLCL. To explore suitable application paradigms, we designed a comprehensive set of experiments applying ZO at varying levels of granularity, from entire modality branches to fine-grained partially trainable units.
However, given the vast diversity of configurations arising from combinations of modality branches and PEFT trainable units, constructing a single, unified theoretical framework is intractable. Consequently, we conduct a comprehensive empirical analysis to investigate both the differential impacts of applying ZO across modality branches and the distinct characteristics of employing it in various trainable layer configurations (such as continuous and interleaved layers), which allows us to achieve optimal results.
}

\section{Study of ZO Optimization in VLCL}
\label{headings}

\subsection{Implementation}

\textbf{Datasets and task construction.}
We evaluate our method on three datasets: CIFAR-100 (CIFAR), Tiny-ImageNet (TinyImg), and ImageNet-R (ImgR). 
For task construction under the CIL paradigm, \newrev{we adopt the IncX configuration (e.g., Inc20 denotes 5 tasks with 20 classes each on CIFAR).} All tasks enforce disjoint class distributions and exclude task-specific identifiers during inference to ensure a rigorous evaluation protocol.

\textbf{Baseline.}
We choose MoE4Adapter~\cite{MoE4Adapters} as the SOTA baseline, which incorporates Mixture of Experts (MoE) into CLIP for VLCL.
To explore PEFT alternatives, we also replaces the MoE modules with Low-Rank Adaptation (LoRA \cite{lora}) modules. For LoRA results, we present a part of them and put the remaining in supplementary material. 

\textbf{Implementation details.}
All experiments employ the CLIP-ViT-B/16 backbone architecture.
The CLIP backbone remains frozen, with only task-specific adapters (MoE or LoRA modules) being trainable. For the ZO-based method, we adopt a more conservative ZO strategy which evaluates multiple candidate updates and selectively applies the one that yields the lowest loss\cite{zeroflow}.
Hyperparameters including perturbation scale \(\varepsilon=0.001\) for ZO gradients and FO/ZO mixing ratio \(\lambda=1\), are validated on the first task and retained for subsequent tasks. 

\subsection{Rethinking ZO Optimization in VLCL}

\textbf{The potential of ZO optimization in VLCL.} \rev{PEFT-based VLCL harnesses trainable units in vision-language model to achieve parameter-efficient adaptation.
Conventional FO optimization method relies on precise gradient descent, hence leading to the attraction by local optima and suboptimal convergence.}
To mitigate these challenges, we first explore integrate full ZO optimization into VLCL. By leveraging its perturbation-based search mechanism, ZO enhances exploration in the parameter space, enabling escape from local optima.
\textit{However, does naively replacing FO with ZO necessarily lead to better performance?}

\begin{table}[t]
    \centering
    \fontsize{8pt}{9pt}\selectfont 
    \setlength{\tabcolsep}{1.2pt} 
    \begin{tabular*}{\columnwidth}{@{\extracolsep{\fill}} l*{8}{c} @{}} 
        \toprule
        \multirow{2}{*}{\textbf{Method}} & 
        \multicolumn{2}{c}{\textbf{CIFAR Inc20}} & \multicolumn{2}{c}{\textbf{CIFAR Inc10}} & \multicolumn{2}{c}{\textbf{TinyImg Inc20}} & \multicolumn{2}{c}{\textbf{ImgR Inc20}} \\
        \cmidrule(lr){2-9} 
         &   Last. & Avg. & Last. & Avg. & Last. & Avg. & Last. & Avg. \\
        \midrule
        Baseline &   80.47 & 86.97 & 77.52 & 85.21 & {52.13} & 60.55 & 65.36 & 71.53 \\
        \hspace{1em} \textit{Du. w/ ZO} &   69.29 & 77.36 & 67.64 & 75.88 & 42.40 & 47.64 & 58.56 & 65.92 \\
        \hspace{1em} \textit{Vis. w/ ZO} &   76.05 & 83.93 & 72.98 & 82.08 & 49.65 & 57.90 & 62.54 & 69.84 \\
        \hspace{1em} \textit{Lan. w/ ZO}&   80.94 & 87.00 & 76.74 & 85.03 & 49.14 & 58.69 & 64.38 & 70.38 \\
        \midrule
        Baseline\dag &  80.44 &87.10 &79.66 &86.34 &51.93 &59.80 &64.34 &71.79 \\
        \hspace{1em} \textit{Du. w/ ZO} &  71.07 &78.37 &69.70 &76.82 &44.53 &52.71 &58.71 &65.97 \\
        \hspace{1em} \textit{Vis. w/ ZO} &  75.86 &83.86 &73.73 &82.25 &49.98 &58.11 &62.70 &69.88 \\
        \hspace{1em} \textit{Lan. w/ ZO} &  77.47 &85.40 &79.63 &87.01 &49.90 &58.92 &64.33 &70.59 \\
        \bottomrule
    \end{tabular*}
    \caption{How ZO optimization affects VLCL in different branches (\texttt{CLIP}). \textit{w/ ZO} denotes the branch (\textit{Du. (Dual)}, \textit{Vis. (Vision)}, or \textit{Lan. (Language)}) where ZO optimization is applied.
The $\dag$ indicates MoE modules in baseline are replaced with LoRA.}
    \label{3_2}
    \vspace{-2ex}
\end{table}

\begin{figure}[t!]
    \centering
    \small
    \begin{minipage}{\columnwidth}
        \begin{subfigure}[b]{0.48\columnwidth}
            \centering
            \includegraphics[width=\textwidth]{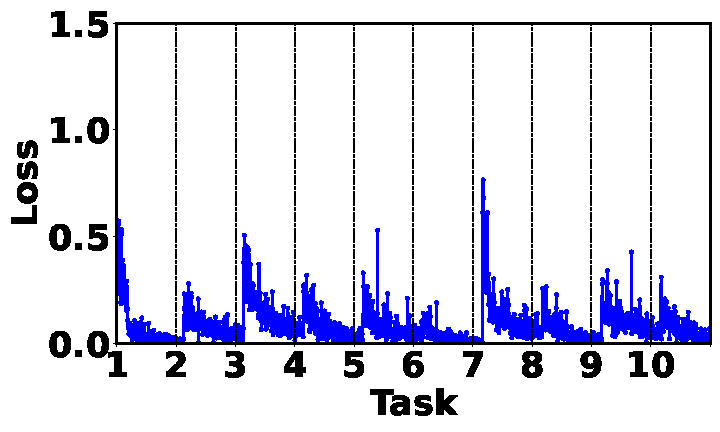}
            \caption{\textit{Dual w/ FO}}
            \label{fig:none_zo_loss}
        \end{subfigure}
        \hfill 
        \begin{subfigure}[b]{0.48\columnwidth}
            \centering
            \includegraphics[width=\textwidth]{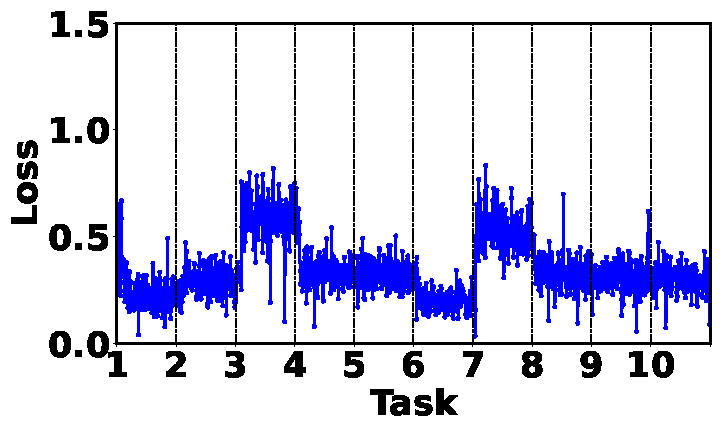}
            \caption{\textit{Dual w/ ZO}}
            \label{fig:dual_zo_loss}
        \end{subfigure}

        \par
        
        \begin{subfigure}[b]{0.48\columnwidth}
            \centering
            \includegraphics[width=\textwidth]{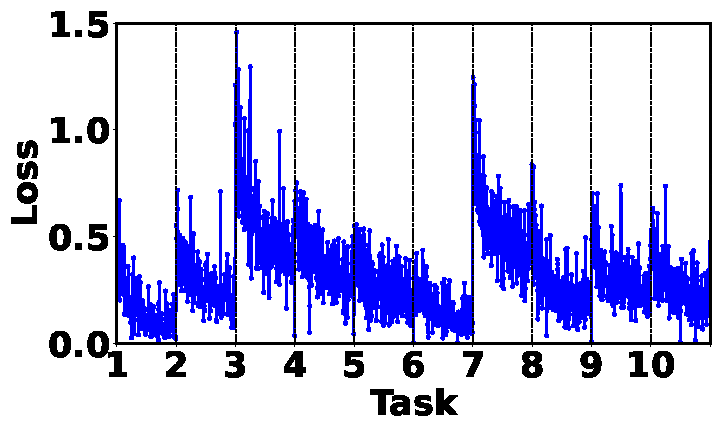}
            \caption{\textit{Vision w/ ZO}}
            \label{fig:Vis_zo_loss}
        \end{subfigure}
        \hfill
        \begin{subfigure}[b]{0.48\columnwidth}
            \centering
            \includegraphics[width=\textwidth]{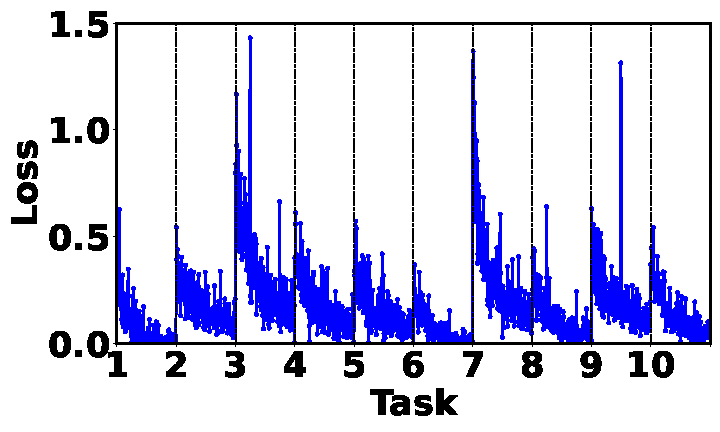}
            \caption{\textit{Language w/ ZO}}
            \label{fig:lan_zo_loss}
        \end{subfigure}
    \end{minipage}
    \caption{How ZO optimization affects loss convergence of VLCL in different branches (\texttt{CLIP}). \textit{w/ ZO} denotes the branch (\textit{Dual}, \textit{Vision}, or \textit{Language}) where ZO applied.}
    \label{fig:zo_comparison}
    \vspace{-3ex}
\end{figure}

\textbf{{Analysis of naive ZO optimization failure in VLCL.}} We attempt to apply ZO into the VLCL,
a straightforward intuition is to replace FO optimizers with ZO methods across both the vision and language branches including all trainable units, and the results are shown in Table~\ref{3_2}. \rev{However, it can be observed when ZO is adopted in both vision and language branch (dual \textit{w/ ZO}), the performances are significantly degraded regardless of MoE or LoRA settings, with Last. and Avg. averagely decreasing by 8.5\% and 9.5\% respectively. It can be attributed to the reason that the variance of ZO destabilizes VLCL training, leading to the optimization process is difficult to converge.}
To verify this point, we plot the loss function trajectories across these experimental settings shown in Figure~\ref{fig:zo_comparison}. The Figure~\ref{fig:none_zo_loss} and~\ref{fig:dual_zo_loss} reveal that dual-branch ZO approach suffers from severe loss oscillations and failed convergence compared to the original FO optimization. From this observation, we conclude that full ZO optimization is fundamentally ill-suited for VLCL, as its inherent gradient estimation fluctuations induce training instability. On the contrary, {FO optimization provides stable gradient directions throughout the training process.} This raises the question: \textit{Can synergistic integration of ZO and FO optimization achieve better performance? }

\textbf{How can ZO optimization be effective in VLCL? Branches, or Layers?} We further investigate a heterogeneous optimization strategy: applying ZO to only one modality branch while retaining FO for the other. 
{From a qualitative perspective, Figure~\ref{fig:Vis_zo_loss} and~\ref{fig:lan_zo_loss} reveal when \rev{ZO is adopted into a single branch (vision or language), the overall loss function trajectory is promising to converge compared with dual branch ZO. We consider that the single-branch FO provides optimization stability, guiding the overall training process maintains consistency. Meanwhile, we can find that the performance obtains significant improvement shown in Table~\ref{3_2}, most of the results are close to the baseline, and even some results outperform baseline. It can be explained that ZO's perturbation-based gradient estimation introduces controlled stochasticity into the optimization process, which probabilistically assists the optimizer in evading suboptimal local minima. Additionally, we find that the performance of language \textit{w/ ZO} generally outperform vision \textit{w/ ZO}, we argue that the optimization stability of language branch is stronger than vision branch, which is verified by the loss trajectories of Figure~\ref{fig:Vis_zo_loss} and~\ref{fig:lan_zo_loss}. These findings suggest a ZO-FO synergy strategy which balances performance exploration and training convergence in VLCL, validating the feasibility of the integration optimization.}

Building on these observations, the effectiveness of coarse-grained integration motivates us to explore the potential of synergistic optimization between ZO and FO methods. We further consider a fine-grained perspective: \textit{Can layer-wise allocation of ZO and FO optimization within modality-specific enhance CL performance?}

\subsection{Why Layers Matter: Triggering Effective Continual Adaptation}

\rev{To explore the performance regarding layer-wise allocation of ZO and FO, four layer-wise ZO patterns were tested: \textit{Hop-\textcolor{gray}{odd}} (adopt ZO in odd layers), \textit{Hop-\textcolor{gray}{even}} (adopt ZO in even layers), \textit{Prefix (\textcolor{gray}{six}}) (adopt ZO in first six layers) and \textit{Suffix (\textcolor{gray}{six}}) (adopt ZO in last six layers)
in dual or single modality branch, with FO used in other layers. The baseline and selection of modality branch is still refer to Table~\ref{3_2}. }

\begin{table}[t]
    \centering
    \fontsize{7.9pt}{9pt}\selectfont 
    \setlength{\tabcolsep}{1pt} 
    \begin{tabular}{lcccccccc}
        \toprule
        \multirow{2}{*}{\textbf{Method}} &
        \multicolumn{2}{c}{\textbf{CIFAR Inc20}} & \multicolumn{2}{c}{\textbf{CIFAR Inc10}} & \multicolumn{2}{c}{\textbf{TinyImg Inc20}} & \multicolumn{2}{c}{\textbf{ImgR Inc20}} \\
        \cmidrule(lr){2-9} 
         &   Last. & Avg. & Last. & Avg. & Last. & Avg. & Last. & Avg. \\
        \midrule
                \textit{Dual w/ ZO}& 69.29 & 77.36 & 67.64 & 75.88 & 42.40 & 47.64 & 58.56 & 65.92 \\
        \hspace{1em} \textit{w/ Hop-\textcolor{gray}{odd}} &{81.64} & {88.11} & {79.22}&{86.98} &51.68 &59.75 &65.24 &71.59 \\
        \hspace{1em} \textit{w/ Hop-\textcolor{gray}{even}}& {81.59} & {88.07} &{79.12} &{86.81} &51.95 &60.53 & 64.99&71.29 \\
        \hspace{1em} \textit{w/ Prefix (\textcolor{gray}{six}})&{79.12} & {85.89}&{76.53} & {84.77}& 50.47&59.34 & 65.01&71.36 \\
        \hspace{1em} \textit{w/ Suffix (\textcolor{gray}{six}})&{80.98} &{87.52} &{78.03} & {86.63}&51.48 &59.33 & 63.10&69.99 \\
        \midrule
        \textit{Vision w/ ZO} & 76.05 & 83.93 & 72.98 & 82.08 & 49.65 & 57.90 & 62.54 & 69.84 \\
        \hspace{1em} \textit{w/ Hop-\textcolor{gray}{odd}} &81.83 &88.36&\textbf{79.39} &86.96&51.56 &60.11 & \textbf{65.68}&72.05 \\
        \hspace{1em} \textit{w/ Hop-\textcolor{gray}{even}}&\textbf{82.41} &88.36 &78.95 &86.74 &\textbf{52.27} &\textbf{60.98} &64.99 & 72.10\\
        \hspace{1em} \textit{w/ Prefix (\textcolor{gray}{six}})&79.34 &86.17 &76.72 &84.85 &50.14 &59.73 &65.25 &71.96 \\
        \hspace{1em}  \textit{w/ Suffix (\textcolor{gray}{six}})&82.21 &88.33 &79.23 &87.03 &51.65 &60.20 &64.25 &70.76 \\
        \midrule
        \textit{Language w/ ZO} & 80.94 & 87.00 & 76.74 & 85.03 & 49.14 & 58.69 & 64.38 & 70.38 \\
        \hspace{1em} \textit{w/ Hop-\textcolor{gray}{odd}}&82.28 & \textbf{88.51} &79.28 &\textbf{87.05} &51.73 &60.59 &65.20 &71.90 \\
        \hspace{1em} \textit{w/ Hop-\textcolor{gray}{even}}&82.17 & 88.45&78.73 &86.27 &52.07 &60.71 &65.06 &\textbf{72.16} \\
        \hspace{1em} \textit{w/ Prefix (\textcolor{gray}{six}})&82.19 &88.51 &79.07 &86.83 &52.02 &60.49 &64.87 &71.64 \\
        \hspace{1em} \textit{w/ Suffix (\textcolor{gray}{six}})&82.19 &88.30 &78.82 &86.71 &51.57 &60.38 &65.17 &72.10 \\
        \bottomrule
    \end{tabular}
    \caption{How ZO optimization affects VLCL through different layers (\texttt{CLIP}). We design four configurations across layers from different branches: \textit{w/ Hop-\textcolor{gray}{odd}} (ZO in odd layers), \textit{w/ Hop-\textcolor{gray}{even}} (ZO in even layers), \textit{w/ Prefix (\textcolor{gray}{six}}) (ZO in first six layers) and \textit{w/ Suffix (\textcolor{gray}{six}}) (ZO in last six layers).}
    \label{3_2_2}
    \vspace{-2ex}
\end{table}

\textbf{Layer-wise ZO unlocks VLCL performance potential.
}
\rev{We then apply layer-wise ZO optimization to both dual and single modality branches, with the remaining layers optimized by FO. 
As shown in Table~\ref{3_2_2}, we observe that layer-wise strategy can provide significant performance improvements compared to applying ZO across all trainable units in branches. In dual \textit{w/ ZO}, the layer-wise ZO optimization averagely improves 9.4\% accuracy on four patterns across all dataset. More interestingly, we observe that certain fine-grained layer-wise ZO patterns can outperform full FO approaches. To investigate the layer-wise effectiveness, we first analyze the performance of applying the collaborative ZO-FO optimization to a uniform layering strategy (with \textit{Hop-\textcolor{gray}{odd}} selected as a representative case). The loss trajectories under different modalities are recorded and presented in Figure~\ref{fig:cross_layer_loss}, it demonstrates a more stable convergence process during training compared to that shown in Figure~\ref{fig:zo_comparison}. We argue that more fine-grained layer-wise strategy further amplifies the respective advantages of ZO and FO in VLCL, FO provides stable gradient directions and ZO's stochastic perturbations help the optimization escape from local minima.

{\textbf{Observation on Layer-wise Heterogeneity in ZO for VLCL.
}}
To gain deeper insights, we start to analyze the impact of different layer-wise settings on performance. Interestingly, we observe that all SOTA results emerge when ZO and FO optimization are applied in an interleaved manner across layers (e.g., on \textit{Hop-\textcolor{gray}{odd}} or \textit{Hop-\textcolor{gray}{even}}). To further investigate this phenomenon, we analyze the gradient behavior under four different layer-wise configurations within the language branch. The corresponding gradient distributions are visualized in Figure~\ref{fig:layeranalisys}. We clearly observe that the gradient variance is significantly lower when ZO and FO are interleaved across layers, compared to configurations where either ZO or FO is applied continuously throughout. We hypothesize that this benefit stems from the functional heterogeneity across layers: shallow layers focus on local features, while deeper layers capture abstract semantics. A uniform optimization method may overlook such diversity, whereas interleaving ZO and FO better aligns with each layer's exploration and stability needs, leading to a more robust optimization process that stabilizes gradient flow while facilitating escape from local minima.}

\begin{figure}[t]
    \centering
    \small
    \begin{minipage}{\columnwidth}
        \begin{subfigure}[b]{0.49\columnwidth}
            \centering
            \includegraphics[width=\textwidth]{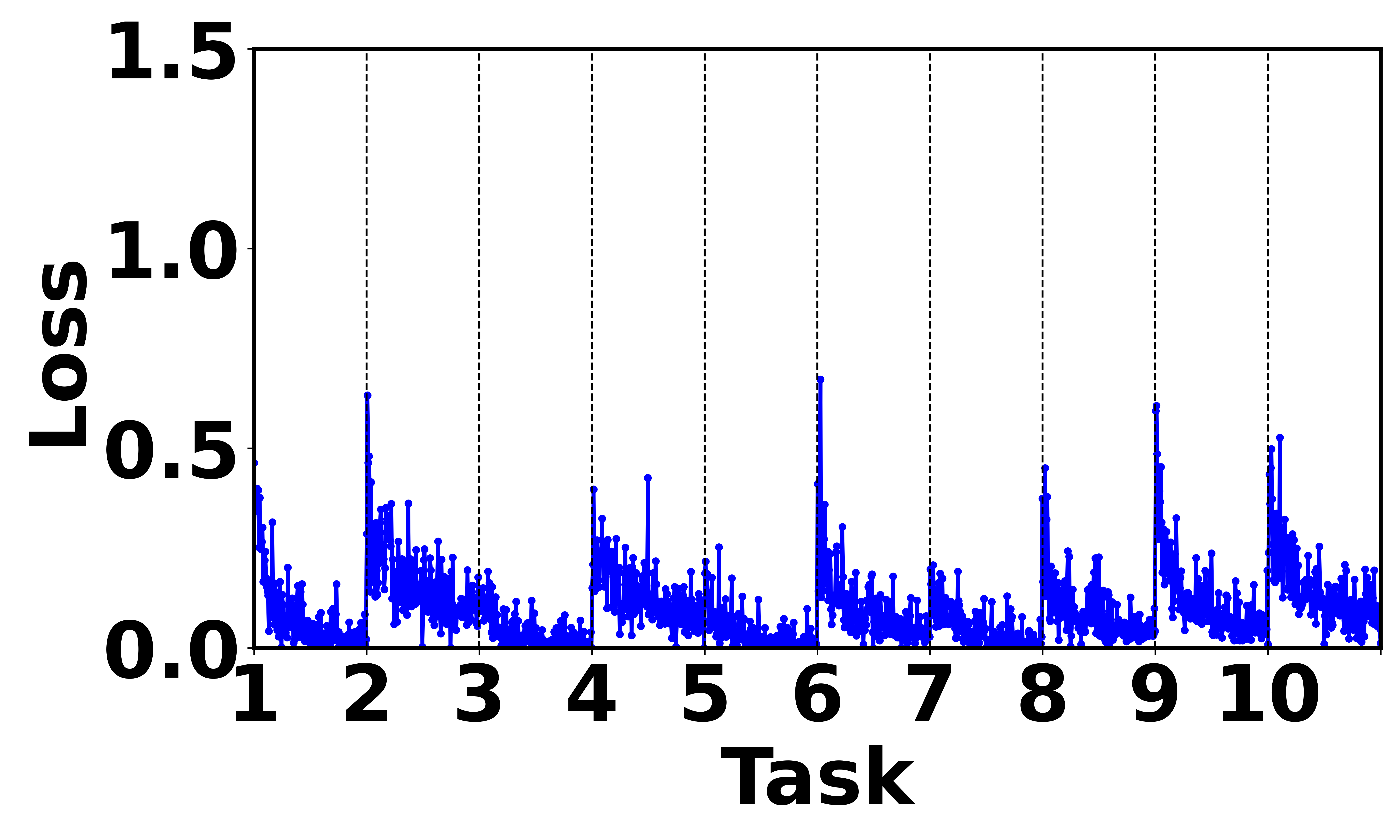}
            \caption{Du. \textit{w/Hop-\textcolor{gray}{odd}}.}
            \label{fig:dualoddloss}
        \end{subfigure}
        \hfill
        \begin{subfigure}[b]{0.49\columnwidth}
            \centering
            \includegraphics[width=\textwidth]{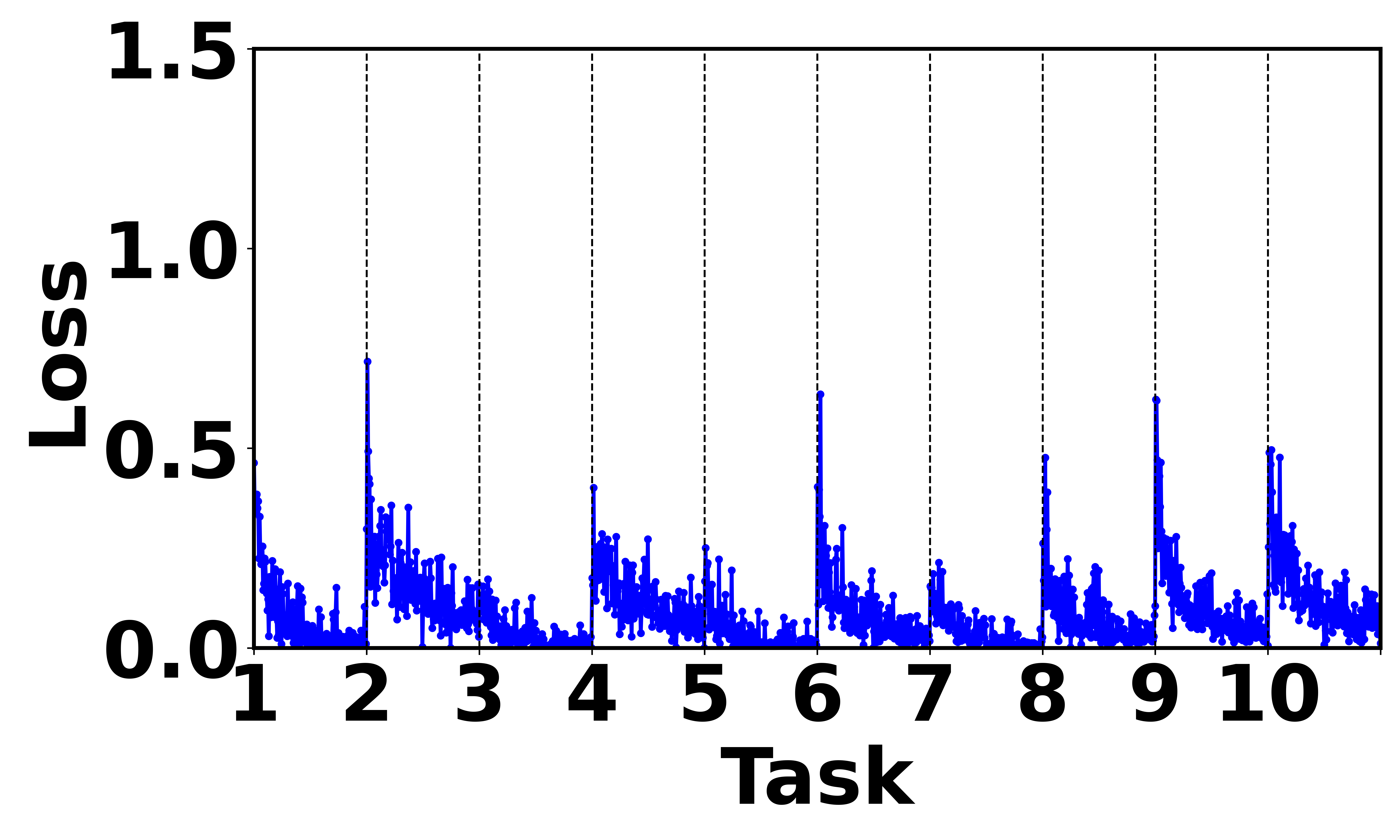}
            \caption{Lan. \textit{w/Hop-\textcolor{gray}{odd}}.}
            \label{fig:languageoddloss}
        \end{subfigure}
    \end{minipage}
    \caption{Analyzing convergence behavior of VLCL in \textit{Hop-\textcolor{gray}{odd}} across Dual (Du.) and Language (Lan.) branches.}
    \label{fig:cross_layer_loss}
    \vspace{-4ex}
\end{figure}

\begin{figure}[hbpt]
    \centering
    \small
    \begin{minipage}{\columnwidth}
        \begin{subfigure}[b]{0.48\columnwidth}
            \centering
            \includegraphics[width=\textwidth]{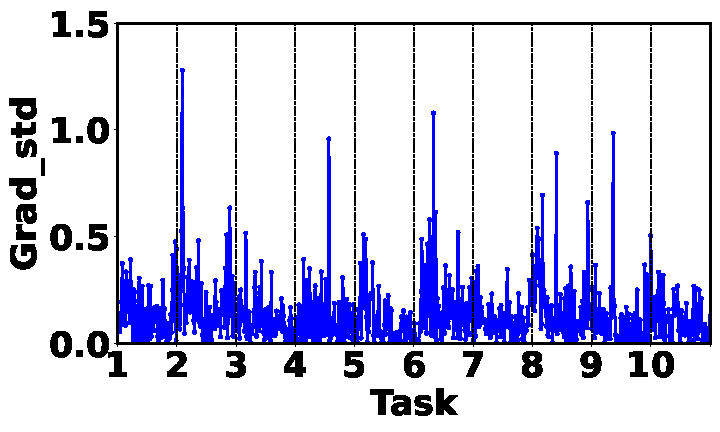}
            \caption{\textit{Prefix (\textcolor{gray}{six}}).}
            \label{fig:front}
        \end{subfigure}
        \hfill 
        \begin{subfigure}[b]{0.48\columnwidth}
            \centering
            \includegraphics[width=\textwidth]{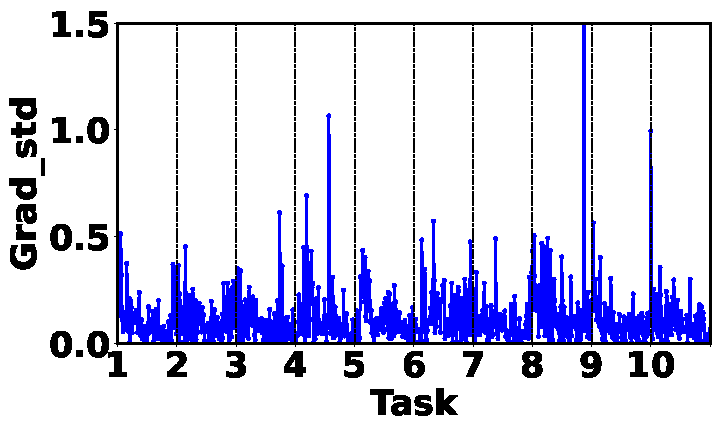}
            \caption{\textit{Suffix (\textcolor{gray}{six}}).}
            \label{fig:back}
        \end{subfigure}
        \par
        \begin{subfigure}[b]{0.48\columnwidth}
            \centering
            \includegraphics[width=\textwidth]{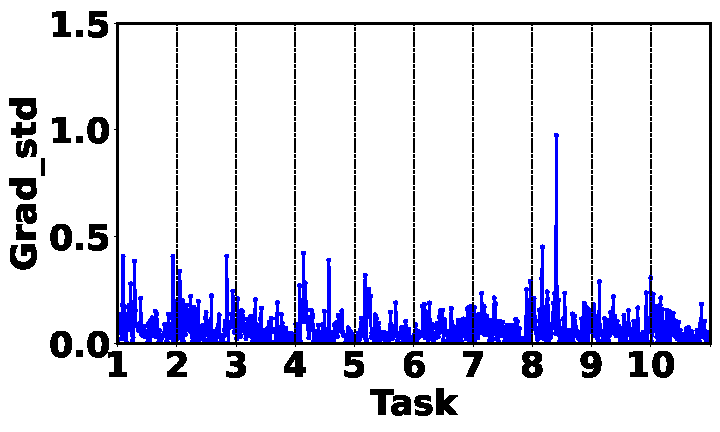}
            \caption{\textit{Hop-\textcolor{gray}{odd}}.}
            \label{fig:odd}
        \end{subfigure}
        \hfill
        \begin{subfigure}[b]{0.48\columnwidth}
            \centering
            \includegraphics[width=\textwidth]{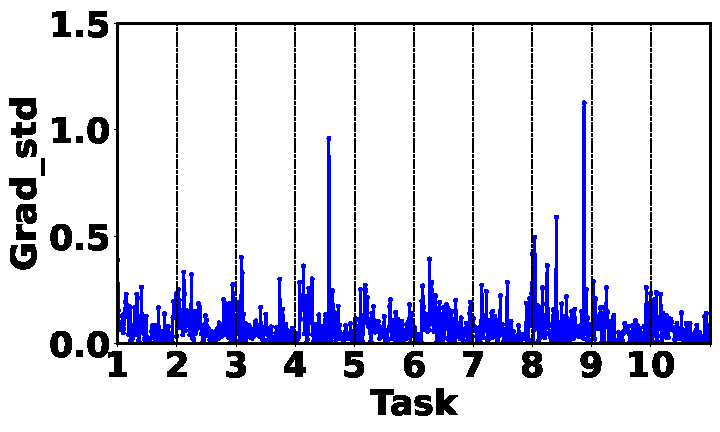}
            \caption{\textit{Hop-\textcolor{gray}{even}}.}
            \label{fig:even}
        \end{subfigure}
    \end{minipage}
    \caption{How ZO optimization affects gradient variance across layers in VLCL.}
    \label{fig:layeranalisys}
    \vspace{-2ex}
\end{figure}

\begin{figure}[b!]
    \centering
    \small
    \begin{minipage}{\columnwidth}
        
        \begin{subfigure}[b]{0.48\columnwidth}
            \centering
        \includegraphics[width=\textwidth]{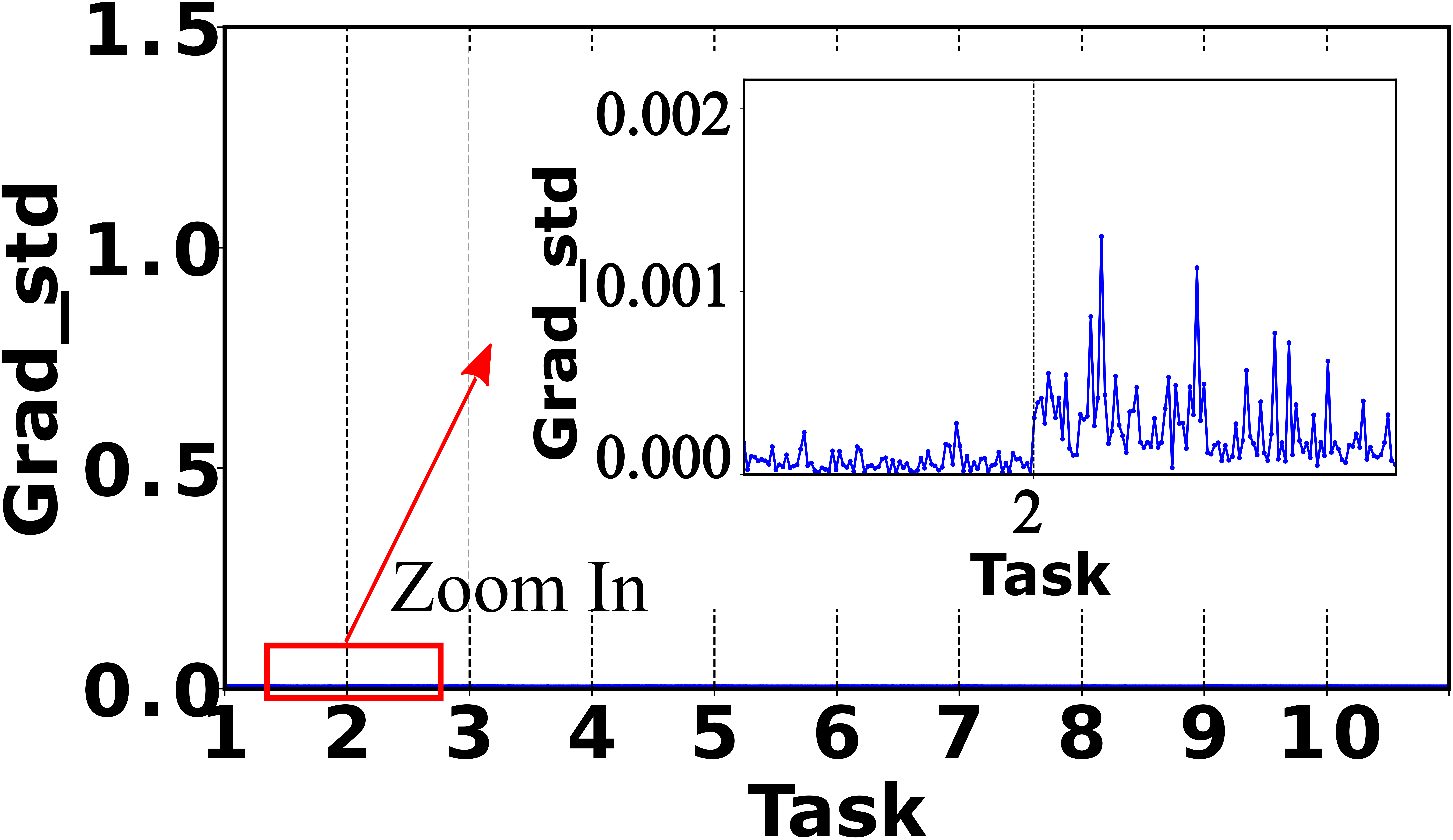}
            \caption{\textit{Dual w/ FO}}
            \label{fig:fooddgrad}
        \end{subfigure}
        \hfill
        \begin{subfigure}[b]{0.48\columnwidth}
            \centering
            \includegraphics[width=\textwidth]{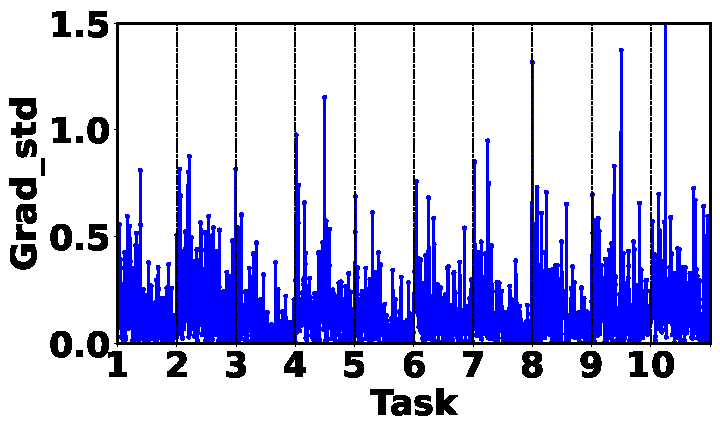}
            \caption{\textit{Dual  w/ ZO}}
            \label{fig:zooddgrad}
        \end{subfigure}
        \hfill 
        \begin{subfigure}[b]{0.48\columnwidth}
            \centering
            \includegraphics[width=\textwidth]{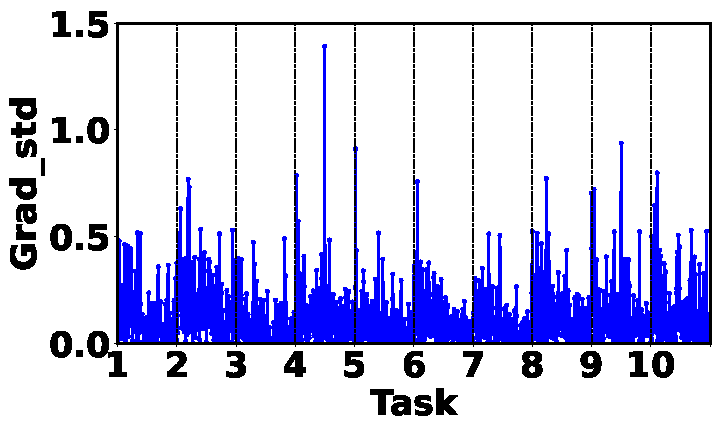}
            \caption{\textit{Vision w/ ZO}}
            \label{fig:visoddgrad}
        \end{subfigure}
        \hfill
        \begin{subfigure}[b]{0.48\columnwidth}
            \centering
            \includegraphics[width=\textwidth]{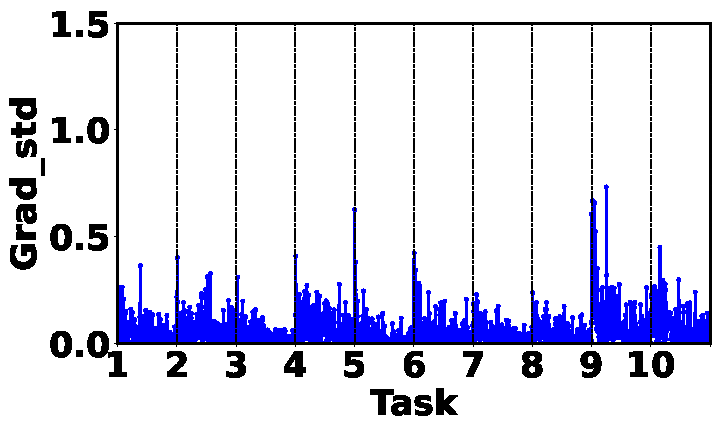}
            \caption{\textit{Language ZO}}
            \label{fig:lanoddgrad}
        \end{subfigure}
    \end{minipage}
    \caption{Analyzing gradient variance of VLCL in \textit{Hop-\textcolor{gray}{odd}} across \textit{Dual}, \textit{Vision}, \textit{Language}.}
    \label{fig:oddgrad}
\end{figure}

\begin{figure}[b!]
    \centering
    
    \small
    \begin{minipage}{\columnwidth}
        
        \begin{subfigure}[b]{0.48\columnwidth}
            \centering
            \includegraphics[width=\textwidth]{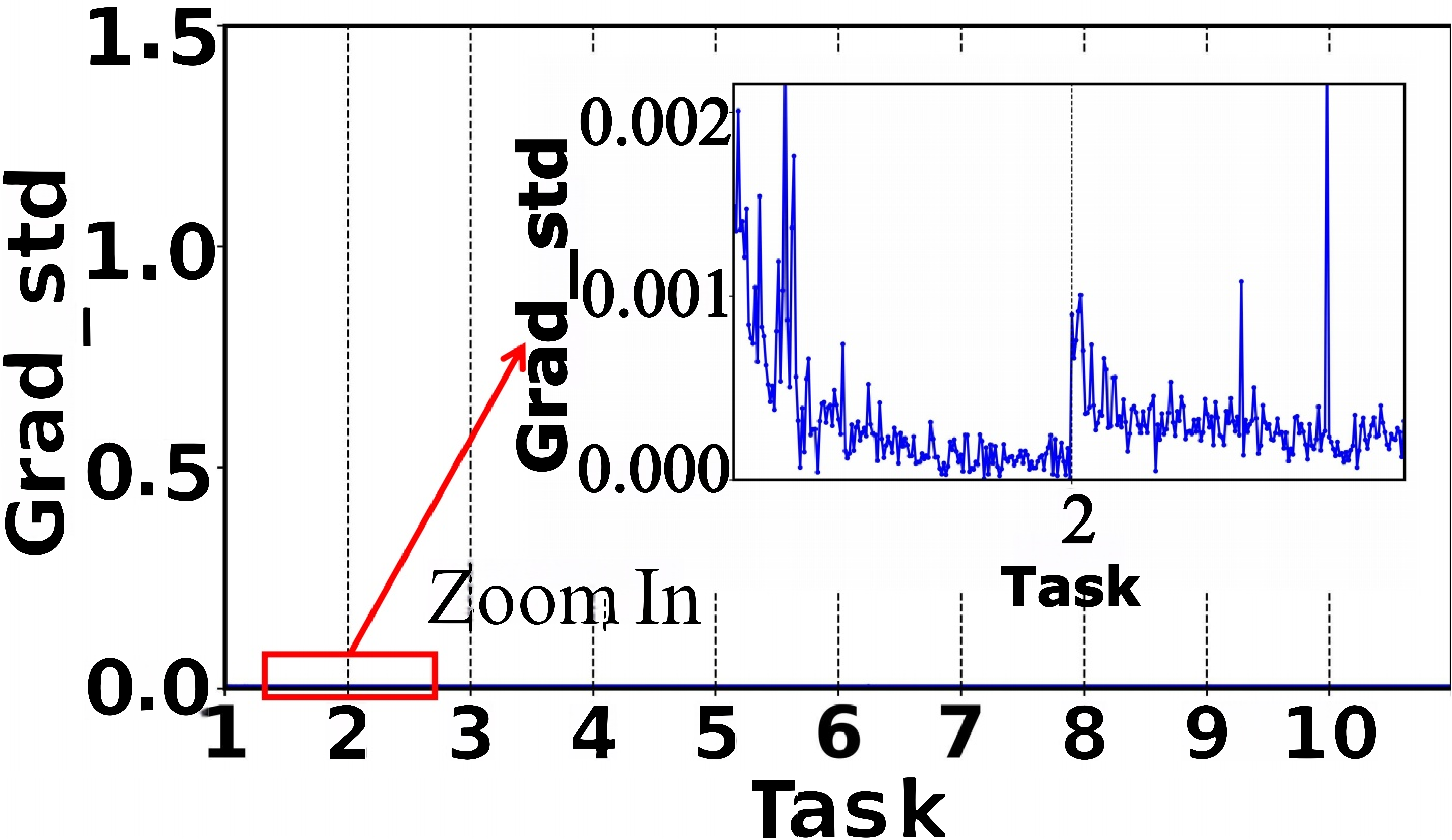}
            \caption{\textit{Dual w/ FO}}
            \label{fig:foevengrad}
        \end{subfigure}
        \hfill
        \begin{subfigure}[b]{0.48\columnwidth}
            \centering
            \includegraphics[width=\textwidth]{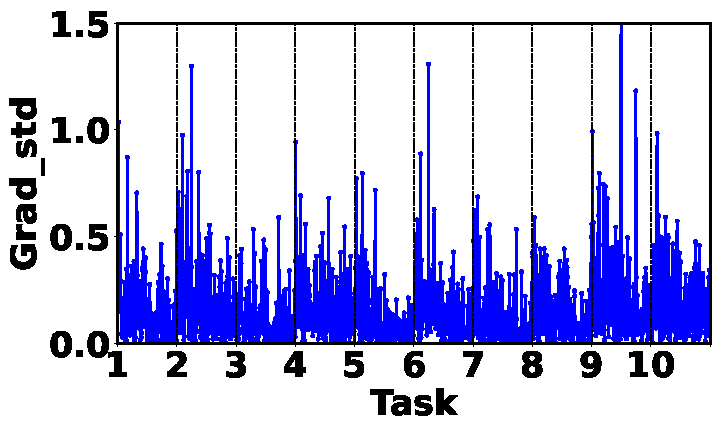}
            \caption{\textit{Dual  w/ ZO}}
            \label{fig:zoevengrad}
        \end{subfigure}
        \par
        \begin{subfigure}[b]{0.48\columnwidth}
            \centering
            \includegraphics[width=\textwidth]{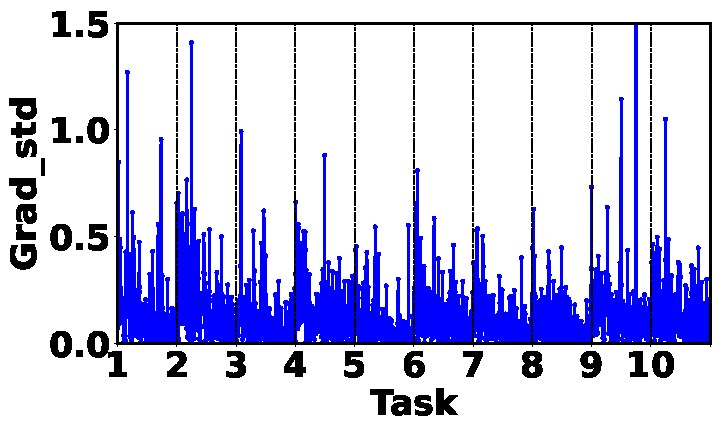}
            \caption{\textit{Vision w/ ZO}}
            \label{fig:visevengrad}
        \end{subfigure}
        \hfill
        \begin{subfigure}[b]{0.48\columnwidth}
            \centering
            \includegraphics[width=\textwidth]{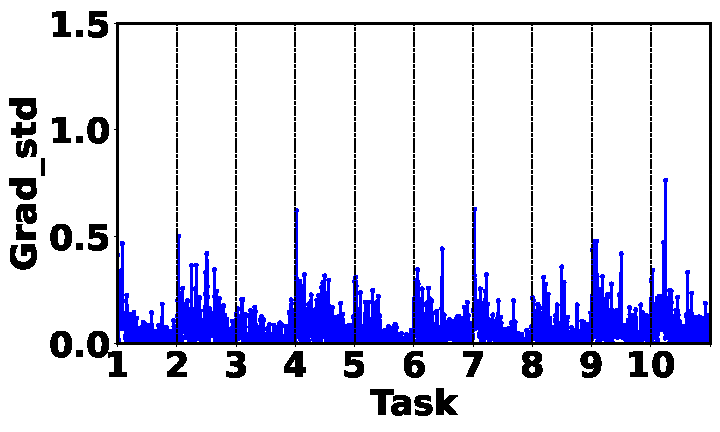}
            \caption{\textit{Language w/ ZO}}
            \label{fig:lanevengrad}
        \end{subfigure}
    \end{minipage}
    \caption{Analyzing gradient variance of VLCL in \textit{Hop-\textcolor{gray}{even}} across \textit{Dual}, \textit{Vision}, \textit{Language}.}
    \label{fig:evengrad}
\end{figure}

\subsection{A New Enhancement from Vision Discrepancy}
\label{zos}

\textbf{Understanding the discrepant behavior of ZO in vision branch.} From Figure~\ref{fig:zo_comparison}, we observe that when ZO is adopted in a single branch, the loss convergence trajectory of visual branch demonstrates significantly inferior performance compared with language branch. To further investigate this phenomenon, we record the gradient variance distribution of different optimization strategies from a layer-wise perspective in VLCL, since layer-wise ZO obtain better performance, and the visualization results are shown in Figure~\ref{fig:oddgrad} and~\ref{fig:evengrad}. It can be observed that dual branch FO optimization exhibit a minimal numerical fluctuations in gradient variance, reflecting robust convergence stability regardless of \textit{Hop-\textcolor{gray}{odd}} or \textit{Hop-\textcolor{gray}{even}} layers setting, while the lack of sufficient fluctuation might lead the optimization process to converge to local optima. In contrast, dual branch ZO induce severe oscillations, causing the optimization trajectory to diverge. When the ZO-FO collaborative mechanism is employed, we interestedly observe that under the same layer wise setting, the gradient variance of the visual branch ZO is also more violent than that of language. This observation motivates a critical inquiry: \textit{Could targeted suppression of ZO-induced perturbations in the visual modality yield performance gains?}

\begin{table}[t]
    \centering
    \small
    \resizebox{\columnwidth}{!}{ 
    \begin{tabular}{lcccccc}
        \toprule
        \multirow{2}{*}{\textbf{Method}} & \multicolumn{2}{c}{\textbf{CIFAR Inc10}}& 
        \multicolumn{2}{c}{\textbf{TinyImg Inc20}}&
        \multicolumn{2}{c}{\textbf{ImgR Inc20}} \\
         &  Last. & Avg. &  Last. & Avg. &  Last. & Avg.\\
        \midrule
        Dual \textit{w/ Hop-\textcolor{gray}{odd}}   & 79.22 & 86.98 &51.68 &59.75 & 65.24 & 71.59 \\
        \textbf{MoZO}  &\textbf{79.36}  & \textbf{87.02}  &\textbf{52.35} &59.75 & \textbf{65.32} & \textbf{71.93}  \\
        \midrule
       Dual \textit{w/ Hop-\textcolor{gray}{even}}   & 79.12 & 86.81 &51.95 &60.53 & 64.99 & 71.29 \\
        \textbf{MoZO}   &\textbf{79.87} & \textbf{87.25} &\textbf{52.46} &\textbf{61.23}  & \textbf{65.80} & \textbf{71.82}  \\
        \bottomrule
    \end{tabular}
    }  
\caption{Effect of MoZO optimization on performance. Dual \textit{w/ Hop-\textcolor{gray}{odd}/\textcolor{gray}{even}} indicates the results of adopting \textit{Hop-\textcolor{gray}{odd}} and \textit{Hop-\textcolor{gray}{even}} layer-wise ZO in dual branch.}
\label{zostratgety}
\vspace{-3ex}
\end{table}

\textbf{Gradient Regularization and Vision Branch Perturbation Control.}
To validate this hypothesis, we propose a MoZO optimization strategy that incorporates gradient regularization during the estimation process, with explicit constraints on perturbations applied to the vision branch to mitigate instability. Specifically, we first introduce a signed gradient transformation to regularize the ZO-estimated gradient $\hat{g}_{ZO}$. This can be expressed as:
\begin{equation}
\tilde{g} = \operatorname{sign}(\hat{g}_{ZO}),
\end{equation}
where $\hat{g}_{ZO}$ denotes the original ZO-estimated gradient, and $\tilde{g}$ is the transformed signed gradient. The $\operatorname{sign}(\cdot)$ function is applied element-wise, defined as:
\begin{equation}
\operatorname{sign}(x_i) =
\begin{cases}
+1. & \text{if } x_i > 0 \\
0. & \text{if } x_i = 0 \\
-1. & \text{if } x_i < 0
\end{cases}
\end{equation}
This transformation retains only the direction information of the gradient, discarding the amplitude information. Furthermore, we implement modality-specific perturbation factors for ZO optimization, assigning a deliberately lower value to the vision branch ($\epsilon_{v}$) compared to the language branch ($\epsilon_{l}$). Hence, the final MoZO update rule can be expressed as:
\begin{equation} \label{eq:mas_zo_unified}
\theta_{m, t+1} =
\begin{cases}
\theta_{m,t} - \eta_t \cdot \tilde{g}(\theta_{m,t}, \epsilon_v\xi_t), & \text{if } m = \text{vision} \\
\theta_{m,t} - \eta_t \cdot \tilde{g}(\theta_{m,t}, \epsilon_l\xi_t), & \text{if } m = \text{language}
\end{cases}
\end{equation}
where $\theta$ is the gradient parameter, $\eta_t$ is the learning rate, and $\xi_t$ represents a perturbation vector.
To verify this strategy, we first record the gradient variance on CIFAR, and plot the comparison shown in Figure~\ref{fig:suppresszo}. It can be observed that our strategy significant reduce the vision branch gradient fluctuations expressed in Figure~\ref{fig:reodd} and~\ref{fig:reeven} (shown in red line), regardless of \textit{Hop-\textcolor{gray}{odd}} or \textit{Hop-\textcolor{gray}{even}}.
We further validate this conjecture through quantitative analysis shown in Table~\ref{zostratgety}, it can be find that stabilizing ZO gradient estimation and reducing vision branch perturbations yields consistent performance gains across CIFAR, TinyImg and ImgR datasets. This method focus on the instability limits of ZO in VLCL and maintains the balance of the optimization process across different modality branches, paving the way to further refines the application strategy of ZO optimization in VLCL.

\begin{figure}[t!]
    \centering
    \small
    \begin{minipage}{\columnwidth}
        
        \begin{subfigure}[b]{0.48\columnwidth}
            \centering
            \includegraphics[width=\textwidth]{figures/visual_grad.png}
            \caption{\textit{Hop-\textcolor{gray}{odd}}}
            \label{fig:rawodd}
        \end{subfigure}
        \hfill
        \begin{subfigure}[b]{0.48\columnwidth}
            \centering
            \includegraphics[width=\textwidth]{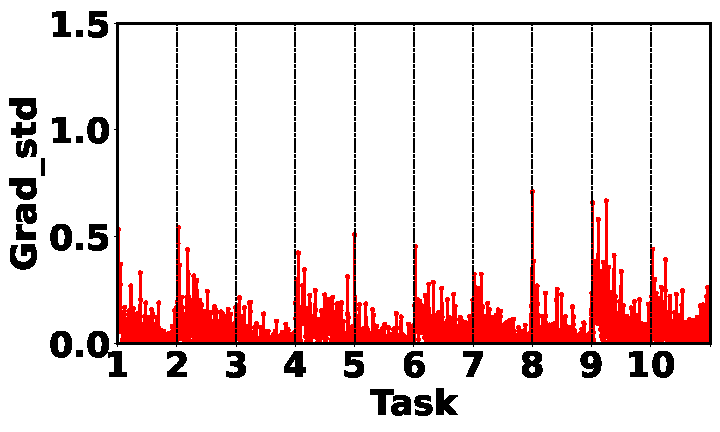}
            \caption{\textit{Hop-\textcolor{gray}{odd}}}
            \label{fig:reodd}
        \end{subfigure}
        \par
        \begin{subfigure}[b]{0.48\columnwidth}
            \centering
            \includegraphics[width=\textwidth]{figures/visual_even.png}
            \caption{\textit{Hop-\textcolor{gray}{even}}}
            \label{fig:raweven}
        \end{subfigure}
        \hfill
        \begin{subfigure}[b]{0.48\columnwidth}
            \centering
            \includegraphics[width=\textwidth]{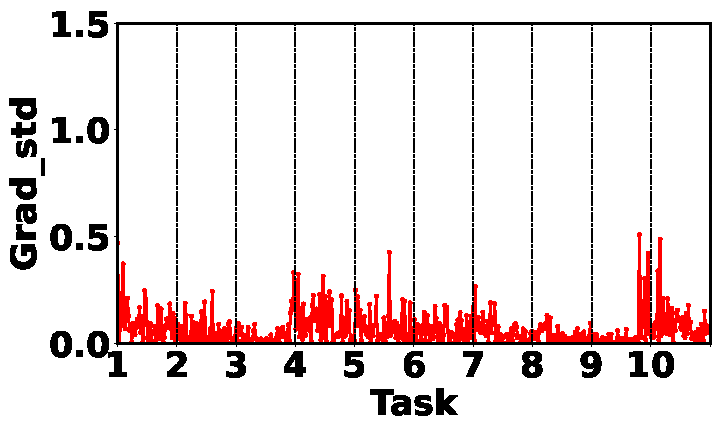}
            \caption{\textit{Hop-\textcolor{gray}{even}}}
            \label{fig:reeven}
        \end{subfigure}
    \end{minipage}
    \caption{Analyzing the effect of MoZO optimization in \textit{Hop-\textcolor{gray}{odd}} vs. \textit{Hop-\textcolor{gray}{even}}. Blue shows original results, while red shows the positive impact of vision discrepancy.}
    \label{fig:suppresszo}
    \vspace{-2ex}
\end{figure}

\section{Ablation, Demonstration and Beyond}

\textbf{Exploring diverse ZO strategies.} Our ablation start to explore other ZO strategies of hybrid ZO-FO collaboration in VLCL. As shown in Table \ref{tab:zo_strategy}, applying ZO to dual branches significantly degrades performance compared to the baseline, confirming the instability caused by excessive gradient oscillations. 
In contrast, single-branch ZO optimization mitigates this issue, with the significant improvement in Last. and Avg. metrics. We then explore the impact of different ZO optimization on VLCL performance. As we default to the conservative ZO strategy in the main analysis, we further conduct more aggressive ZO variants. Specifically, we examine a naive ZO approach—referred to as ZO* which performs a single gradient estimation and directly updates the parameters without any loss-based validation. 
On top of this, we introduce a variant named Sign, which incorporates a signed gradient transformation to regulate the magnitude of the estimated gradients. This design aims to mitigate the instability caused by ZO estimates while preserving directional information. It can be observed that the aggressive ZO* optimization degrades performance, as it relies on a single gradient estimation without validating the update direction, making it more sensitive to the fluctuations in the loss landscape compared with conservative ZO.
\begin{table}[t!]
    \centering
    \small
    \begin{tabular}{lccc}
        \toprule
        \multirow{2}{*}{\textbf{Method}} & \multirow{2}{*}{\textbf{Strategy}} & \multicolumn{2}{c}{\textbf{CIFAR Inc10}} \\
         &   & Last. & Avg. \\
        \midrule

        MoE4Adapter &    &  &   \\
        \hspace{1em} \textit{Dual w/ ZO} &  ZO* & 66.67 & 75.08 \\
        \hspace{1em} \textit{Vision w/ ZO} & ZO*   & 72.32 & 81.64  \\
        \hspace{1em} \textit{Language w/ ZO} & ZO*  & \textbf{76.19} & \textbf{84.58}  \\
        \midrule
        \hspace{1em} \textit{Dual w/ ZO} &  Sign  & 66.68 & 77.20 \\
        \hspace{1em} \textit{Vision w/ ZO} & Sign  & 73.34 & 83.47  \\
        \hspace{1em} \textit{Language w/ ZO} & Sign   & \textbf{78.52} & \textbf{85.91}  \\
        \bottomrule
    \end{tabular}
        \captionof{table}{Behavioral consistency across diverse ZO strategies. * indicates a naive ZO strategy, Sign indicates a ZO strategy using the gradient transformation method.}
        \label{tab:zo_strategy}
    \vspace{-1ex}
\end{table} 
Notably, adopting Sign-based gradient estimation enhances performance across all configurations, suggesting that enforcing gradient amplitude consistency in ZO perturbations helps stabilize the optimization trajectory and improves convergence. This observation aligns with our proposed ZO optimization strategy in Section~\ref{zos}, which advocates for gradient magnitude control as a means to suppress fluctuation updates and maintain optimization balance across modalities.

\begin{figure*}[t]
    \centering
    \small
    \begin{minipage}{\textwidth}
        \begin{subfigure}[b]{0.245\textwidth}
            \centering
            \includegraphics[width=\textwidth]{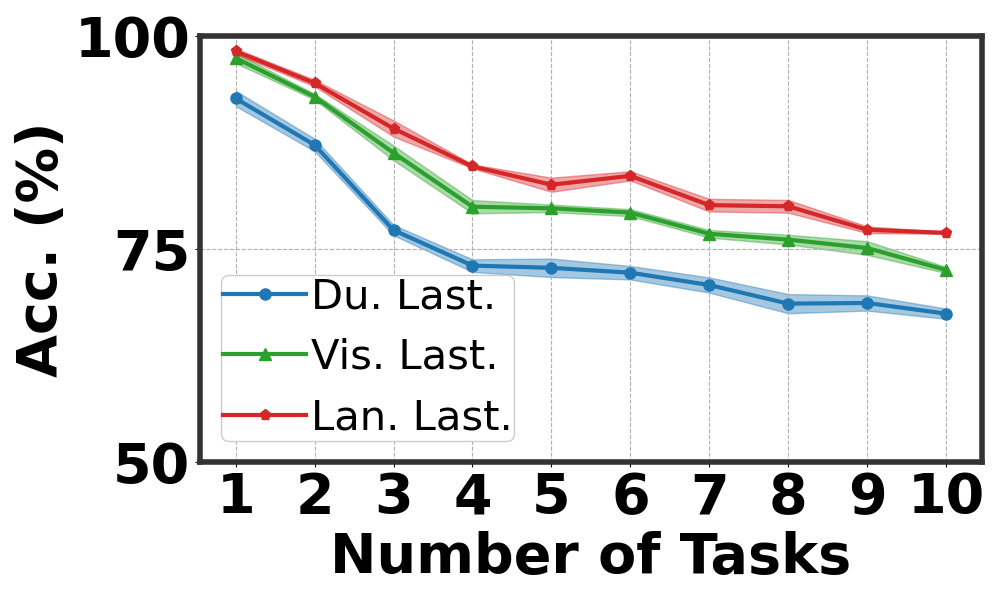}
            \caption{Last. of CIFAR.}
            \label{fig:cifarlast}
        \end{subfigure}
        \hfill 
        \begin{subfigure}[b]{0.245\textwidth}
            \centering
            \hfill
        \includegraphics[width=\textwidth]{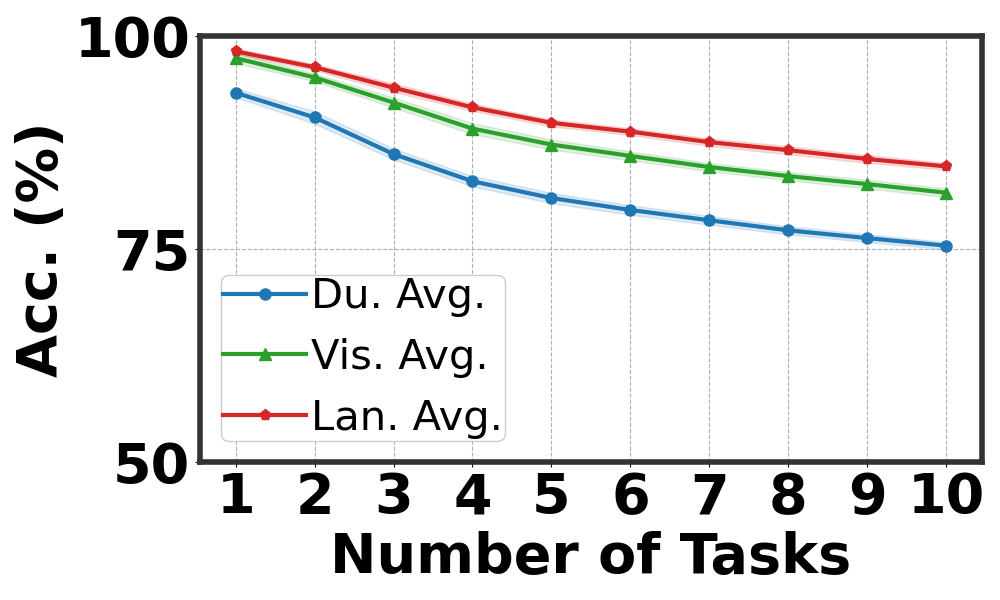}
            \caption{Avg. of CIFAR.}
            \label{fig:cifaravg}
        \end{subfigure}
        \hfill
        \begin{subfigure}[b]{0.245\textwidth}
            \centering
            \includegraphics[width=\textwidth]{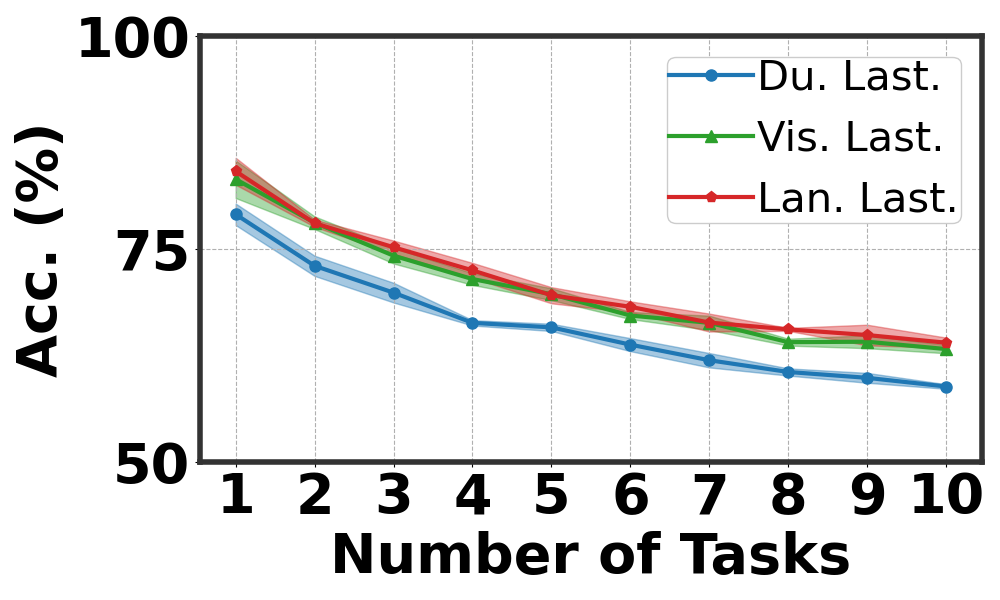}
            \caption{Last. of ImgR.}
            \label{fig:imagerlast}
        \end{subfigure}
        \hfill 
        \begin{subfigure}[b]{0.245\textwidth}
            \centering
        \includegraphics[width=\textwidth]{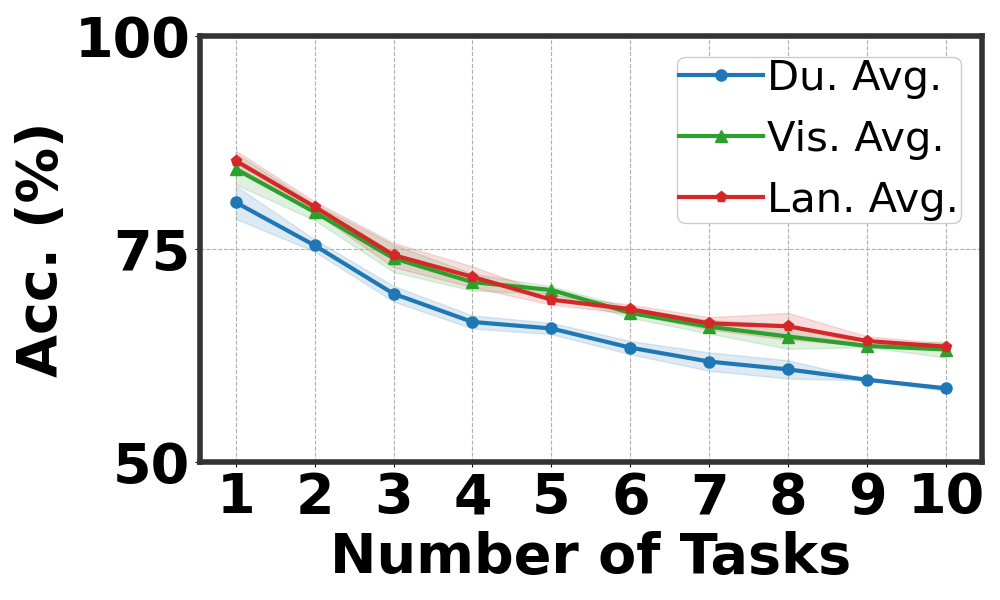}
            \caption{Avg. of ImgR.}
            \label{fig:imageravg}
        \end{subfigure}
    \end{minipage}
    \caption{Significance analysis of performance across \textit{Dual (Du.)}, \textit{Vision (Vis.)}, \textit{Language (Lan.)}.}
    \label{fig:AverageIncremental}
    \vspace{-3ex}
\end{figure*}

\textbf{Significance analysis.} We then explore the average incremental performance among five runs on two dataset (CIFAR and ImgR), with three different ZO configurations (dual branch, vision branch and language branch), recording the accuracy of Last. and Avg. shown in Figure~\ref{fig:AverageIncremental}. We observe that ZO optimization yields smaller performance variance on the CIFAR dataset than on ImgR across different configurations, regardless of Last. and Avg. metrics. One possible explanation is that ImgR derived from the larger-scale ImageNet, contains more fine-grained and semantically diverse categories than CIFAR. This increased label complexity leads to a more rugged and high-dimensional loss landscape, which makes ZO gradient estimation on stochastic perturbations harder to stabilize. Additionally, from the perspective of different ZO configurations, applying ZO optimization to the language branch consistently yields the better performance. We hypothesize that this is because the language branch typically operates on lower-dimensional tensors, making it less susceptible to the instability introduced by random perturbations in ZO gradient estimation. In contrast, the higher-dimensional vision branch benefits more from FO gradients, which provide precise gradient optimization direction. The combination of FO-dominated updates in the vision branch and ZO exploration in the language branch may facilitate better escape from local minima and lead to more robust convergence.

\begin{table}[htbp]
    \centering
    \small
    \begin{tabular}{lccc}
        \toprule
        \multirow{1}{*}{\textbf{ Method }} & \multirow{1}{*}{\textbf{MoE4Adapter}} & \multirow{1}{*}{\textbf{MoE4Adapter\dag}}\\
        \midrule
        Baseline &  $\sim$19.96GB  & $\sim$15.11GB \\
        \textit{Dual w/ ZO} &  $\sim$2.17GB $\downarrow$ & $\sim$1.73GB $\downarrow$\\
        \textit{Vision w/ ZO} & $\sim$6.93GB $\downarrow$   & $\sim$5.71GB $\downarrow$ \\
        \textit{Language w/ ZO} & $\sim$12.39GB $\downarrow$  & $\sim$11.09GB $\downarrow$ \\
        \bottomrule
    \end{tabular}
    \captionof{table}{Comparison of GPU memory usage between different ZO settings across branches (\textbf{\texttt{CLIP})}.}
    \label{tab:memory}
    \vspace{-3ex}
\end{table}

\textbf{Quantifying memory efficiency.} 
We record the memory usage across different ZO settings, as shown in Table~\ref{tab:memory}.
Compared to applying FO optimization in the dual branch baseline, using ZO optimization leads to a 89.1\% reduction in memory consumption, due to the elimination of gradient backpropagation and storage overhead. 
When applying ZO optimization to a single branch, memory consumption is reduced by 65.3\% on the visual branch and 37.9\% on the language branch, respectively. The greater memory reduction on the visual branch is attributed to its higher input dimensionality, as it processes image data. Applying ZO in this context helps to alleviate memory pressure by avoiding storage of high-dimensional gradients. For the results in LoRA architecture, it inherently requires fewer trainable units to perform parameters update, resulting in overall lower memory consumption. Overall, these findings highlight ZO's inherent advantage in significantly reducing memory consumption, especially in high-dimensional settings, making it a highly practical and efficient optimization alternative for resource-constrained VLCL scenarios where memory efficiency is critical.

\section{Related Work}

\textbf{Continual learning of VLM.} Recent advances in CL have witnessed the emergence of VLMs as promising solutions to mitigate catastrophic forgetting through their generalized multimodal representations.
Pioneering studies \cite{lwf-vr, clip_cl2} demonstrate that pretrained VLMs like CLIP \cite{clip} inherently possess remarkable continual learning capabilities even without fine-tuning.
PROOF \cite{lwf-vr} further enhances CL robustness by integrating multimodal cues with adaptive mapping strategies.
Nevertheless, conventional VLM-based approaches predominantly emphasize task-specific feature acquisition for new domains \cite{ralf}, inadvertently compromising the integrity of previously learned representations and leading to progressive performance degradation.
To address this limitation, recent efforts adopt PEFT techniques \cite{peft1, peft2} that selectively update lightweight modules while maintaining frozen backbone parameters. 
MoE4Adapters \cite{MoE4Adapters} introduces a mixture-of-experts (MoE) architecture, enabling task-specific feature specialization without cross-task interference and achieving state-of-the-art performance.
However, these methods universally rely on FO optimization, inherently restricting their capacity to explore optimal parameter trajectories during optimization.

\textbf{Optimization for continual learning.} From an optimization view, existing CL methods predominantly focus on reconciling the stability-plasticity dilemma~\cite{staplast} through gradient technology~\cite{unigrad}.
Orthogonal parameter updates \cite{grad1,grad2,grad3,grad4} and sharpness-aware minimization \cite{flat1,flat2,flat3} represent two mainstream directions, which respectively aim to decouple task-specific gradients and converge to flat loss minima for improved generalization.
While these strategies enhance optimization stability, they often inadequately address the exploration-exploitation trade-off, as deterministic gradient descent trajectories tend to converge to suboptimal local minima with limited perturbation resilience.
Emerging ZO optimization\cite{zo1,derivative} presents a paradigm shift by employing gradient-free stochastic perturbations to estimate descent directions. This approach offers dual advantages: 1) inherent stochasticity facilitates escape from local optima through controlled parameter space exploration, and 2) elimination of backward propagation reduces memory overhead by avoiding gradient matrix storage.

\textbf{Our work.} This work investigates how ZO optimization can be effectively adapted to VLCL, addressing catastrophic forgetting by harmonizing the inter-modal asymmetry of VLMs with the stochastic perturbations of ZO optimization.

\section{Conclusion}
\rev{In this work, we systematically investigate ZO optimization into VLCL and propose a novel hybrid ZO-FO optimization paradigm. Through extensive empirical analysis, we identify two critical challenges of applying ZO in VLCL: destabilized training caused by excessive gradient variance and modality-specific optimization discrepancies. To address these, we first demonstrate that selectively applying ZO to specific branches (vision or language) while retaining FO optimization in others significantly outperforms naive full-ZO approaches. Building on this, we further propose a layer-wise collaborative strategy that interleaves ZO and FO across network layers, achieving state-of-the-art performance by harmonizing stochastic exploration with deterministic refinement.}

\textbf{Limitation.} This work currently focuses on CLIP-based vision-language modalities. The generalization of ZO-FO collaboration to other VLMs (e.g., multimodal transformers with audio or video inputs) remains unexplored, particularly in scenarios where modalities exhibit heterogeneous feature distributions or temporal dependencies. 

\section{Acknowledgements}
This work is supported in part by the National Natural Science Foundation of China (62192783), Young Elite Scientists Sponsorship Program by CAST (2023QNRC001), and China Scholarship Council with award number 202406240114.

\bibliography{reference}

\appendix

\clearpage
\begin{center}
    \LARGE \textbf{Supplementary Material}
\end{center}

\renewcommand{\thesection}{\Alph{section}}

In the main paper, we analyze the application of ZO optimization in VLCL, covering both branch-wise and layer-wise settings. Extensive evaluations are conducted based on the MoE-based setting, and further preliminary explorations are carried out under the LoRA-based setting. In the supplementary, we provide additional results to examine the consistency of our observations under the LoRA setting. This material is divided into two sections:

\begin{itemize}
    \item Investigating the performance of layer-wise allocation of ZO and FO optimization under the LoRA setting, building upon the performance trends and layer heterogeneity analyses presented in the main paper.

    \item Exploring the vision branch discrepancy of ZO optimization and evaluates the proposed method from the main paper within the LoRA setting.
\end{itemize}

\section{Investigating Layer-wise ZO optimization for VLCL under LoRA Setting}
\label{seca}

We give the performance of layer-wise ZO optimization with both dual and single modality branches under LoRA setting shown in Table~\ref{lora3_2_2}. Specifically, we choose two layer configurations: \textit{Hop-\textcolor{gray}{odd}} and \textit{Suffix (\textcolor{gray}{six}}). It can be observed that the performance trends are consistent with those reported in the main paper, demonstrating the layer-wise ZO optimization remains effective under the LoRA setting. We also show the convergence behavior of \textit{Hop-\textcolor{gray}{odd}} on ImgR dataset in Figure~\ref{fig:lora_cross_layer_loss}, and find that the training process remains consistently stable throughout.

\begin{table}[H]
    \centering
    \resizebox{\linewidth}{!}{ 
    \begin{tabular}{lcccccccc}
        \toprule
        \multirow{2}{*}{\textbf{Method}} & 
        \multicolumn{2}{c}{\textbf{CIFAR Inc20}} & \multicolumn{2}{c}{\textbf{CIFAR Inc10}} & \multicolumn{2}{c}{\textbf{TinyImg Inc20}} & \multicolumn{2}{c}{\textbf{ImgR Inc20}} \\
        \cmidrule(lr){2-9} 
         &   Last. & Avg. & Last. & Avg. & Last. & Avg. & Last. & Avg. \\
        \midrule
                \textit{Dual w/ ZO}& 71.07&78.37&69.70 &76.82 &44.53&52.71&58.71 &65.97 \\
        \hspace{1em} \textit{w/ Hop-\textcolor{gray}{odd}} &79.36& 86.93& 78.76&86.14&52.04&60.47&64.90 &71.62 \\
        \hspace{1em} \textit{w/ Suffix (\textcolor{gray}{six}})&79.38&86.55&77.88& 85.67&51.56&60.17& 63.15&70.34 \\
        \midrule
        \textit{Vision w/ ZO} & 75.86&83.86&73.73 &82.25 &49.98&58.11&62.70 &69.88 \\
        \hspace{1em} \textit{w/ Hop-\textcolor{gray}{odd}} &\textbf{79.92}&\textbf{87.13}&79.16&86.53&51.94&60.35& 65.16&\textbf{71.80} \\
        
        \hspace{1em}  \textit{w/ Suffix (\textcolor{gray}{six}})&79.60&86.79&78.96&86.35&52.27&60.75&64.34 &70.99 \\
        \midrule
        \textit{Language w/ ZO} & 77.47&85.40&79.63 &87.01 &49.90&58.92&64.33 &70.59 \\
        \hspace{1em} \textit{w/ Hop-\textcolor{gray}{odd}}&79.82& 86.90&\textbf{79.88}&\textbf{87.38}&\textbf{52.32}&\textbf{61.06}&\textbf{65.59} &71.70 \\
        \hspace{1em} \textit{w/ Suffix (\textcolor{gray}{six}})&79.68&86.94&78.46&86.13&52.22&60.61&65.51 &71.67 \\
        \bottomrule
    \end{tabular}
    }
    \caption{How ZO optimization affects VLCL through different layers (\texttt{CLIP}) under LoRA based setting. We choose two configurations across layers from different branches: \textit{w/ Hop-\textcolor{gray}{odd}} and \textit{w/ Suffix (\textcolor{gray}{six}}).}
    \label{lora3_2_2}
\end{table}

\begin{figure}[hbpt]
    \centering
    \small
    \begin{minipage}{\columnwidth}
        \begin{subfigure}[b]{0.31\columnwidth}
            \centering
            \includegraphics[width=\textwidth]{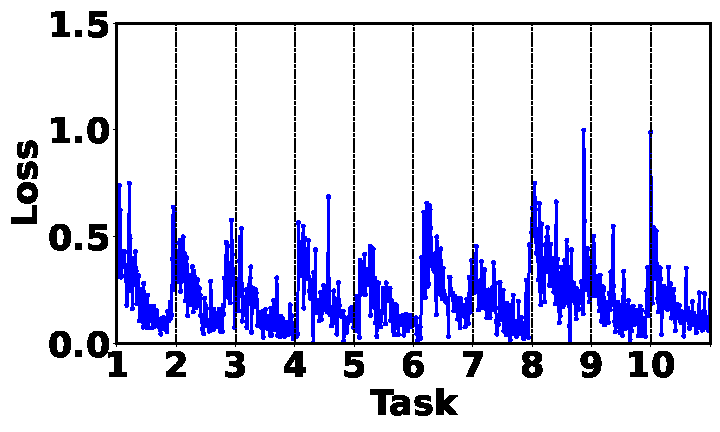}
            \caption{Du. \textit{w/Hop-\textcolor{gray}{odd}}.}
            \label{fig:loradualoddloss}
        \end{subfigure}
        \begin{subfigure}[b]{0.31\columnwidth}
            \centering
            \includegraphics[width=\textwidth]{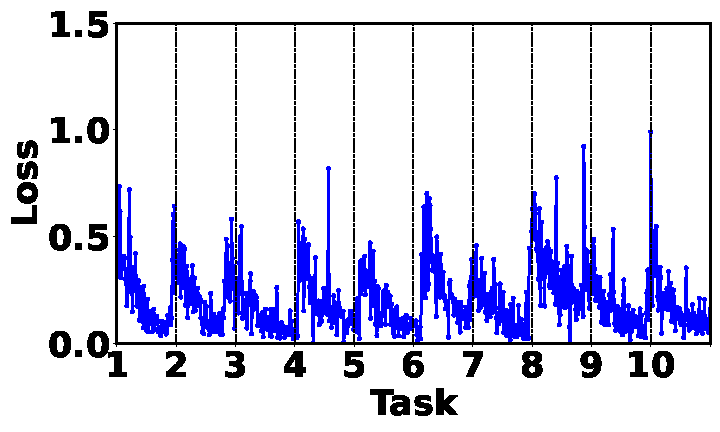}
            \caption{Vis. \textit{w/Hop-\textcolor{gray}{odd}}.}
            \label{fig:loravisionoddloss}
        \end{subfigure}
        \begin{subfigure}[b]{0.31\columnwidth}
            \centering
            \includegraphics[width=\textwidth]{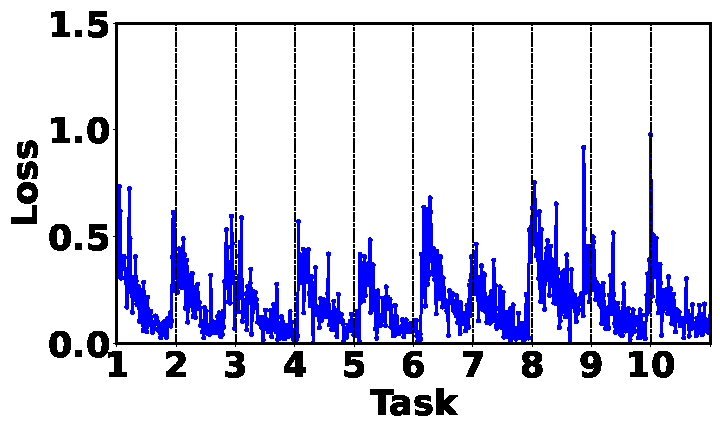}
            \caption{Lan. \textit{w/Hop-\textcolor{gray}{odd}}.}
            \label{fig:loralanguageoddloss}
        \end{subfigure}
    \end{minipage}
    \caption{Analyzing convergence behavior of VLCL in \textit{Hop-\textcolor{gray}{odd}} across Dual (Du.), Vision (Vis.), Language (Lan.) under LoRA setting.}
    \label{fig:lora_cross_layer_loss}
\end{figure}

Additionally, we similarly observe that interleaved layer-wise ZO manner (\textit{Hop-\textcolor{gray}{odd}}) yields superior performance compared with consecutive layers under the LoRA setting. To better understand the results, we separately record the gradient distributions of the \textit{Hop-\textcolor{gray}{odd}} and \textit{Hop-\textcolor{gray}{suf}} layer-wise strategies across the dual, vision, and language branches on the ImgR dataset. Figure~\ref{fig:lora_suf_odd_loss} clearly reveals that interleaved layer-wise ZO optimization yields more stable gradient behavior during training, suggesting interleaving of ZO and FO better meet each layer's needs for gradient exploration and stability, which aligns with the hypothesis proposed in the main paper.

\begin{figure}[hbpt]
    \centering
    \small
    \begin{minipage}{\columnwidth}
        \begin{subfigure}[b]{0.31\columnwidth}
            \centering
            \includegraphics[width=\textwidth]{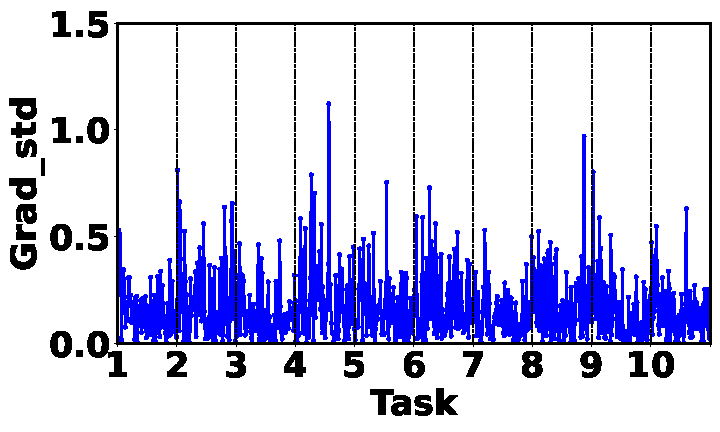}
            \caption{Du. \textit{w/Hop-\textcolor{gray}{suf}}.}
            \label{fig:loradualsuf}
        \end{subfigure}
        \begin{subfigure}[b]{0.31\columnwidth}
            \centering
            \includegraphics[width=\textwidth]{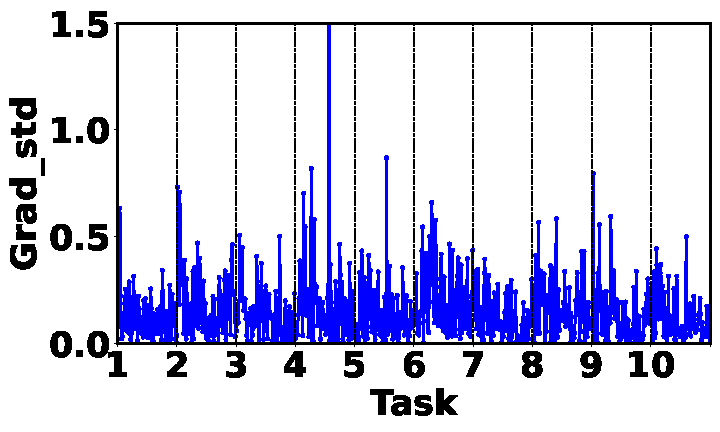}
            \caption{Vis. \textit{w/Hop-\textcolor{gray}{suf}}.}
            \label{fig:loravisionsuf}
        \end{subfigure}
        \begin{subfigure}[b]{0.31\columnwidth}
            \centering
            \includegraphics[width=\textwidth]{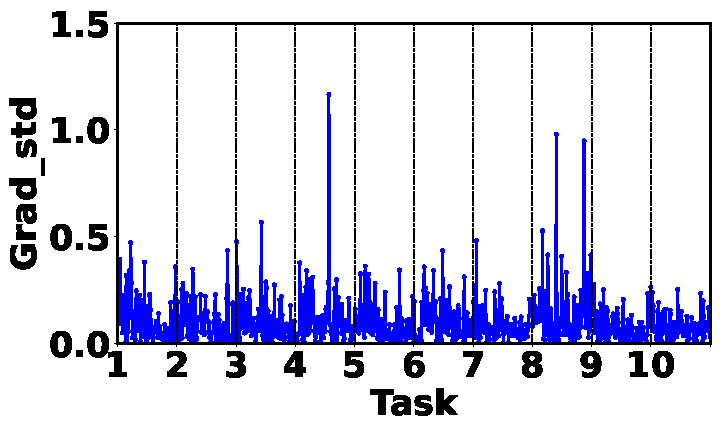}
            \caption{Lan. \textit{w/Hop-\textcolor{gray}{suf}}.}
            \label{fig:loralanguagesuf}
        \end{subfigure}
        \\
        \begin{subfigure}[b]{0.31\columnwidth}
            \centering
            \includegraphics[width=\textwidth]{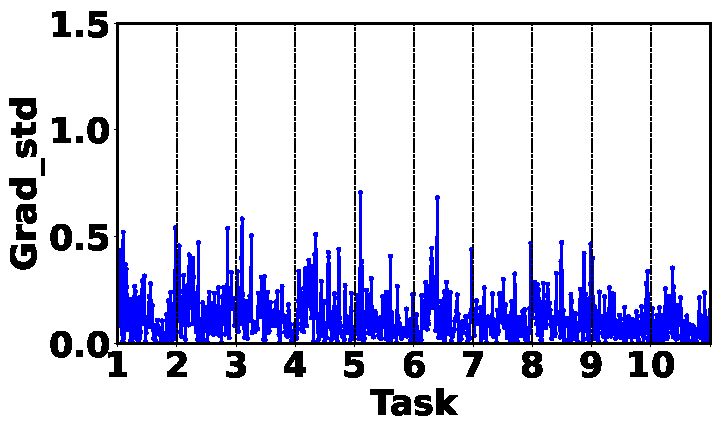}
            \caption{Du. \textit{w/Hop-\textcolor{gray}{odd}}.}
            \label{fig:loradualodd}
        \end{subfigure}
        \begin{subfigure}[b]{0.31\columnwidth}
            \centering
            \includegraphics[width=\textwidth]{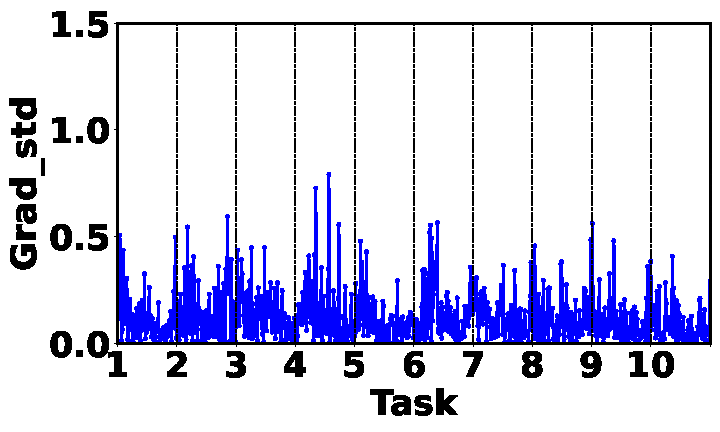}
            \caption{Vis. \textit{w/Hop-\textcolor{gray}{odd}}.}
            \label{fig:loravisionodd}
        \end{subfigure}
        \begin{subfigure}[b]{0.31\columnwidth}
            \centering
            \includegraphics[width=\textwidth]{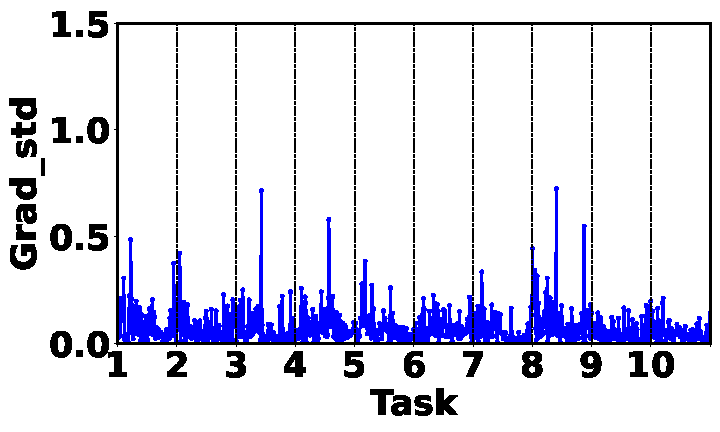}
            \caption{Lan. \textit{w/Hop-\textcolor{gray}{odd}}.}
            \label{fig:loralanguageodd}
        \end{subfigure}
    \end{minipage}
    \caption{How ZO optimization affects gradient variance across layers in Dual (Du.), Vision (Vis.), Language (Lan.) under LoRA setting.}
    \label{fig:lora_suf_odd_loss}
\end{figure}

\begin{table}[b!]
    \centering
    \resizebox{\columnwidth}{!}{ 
    \begin{tabular}{ccccccc}
        \toprule
        \multirow{2}{*}{\textbf{Method}} & \multicolumn{2}{c}{\textbf{CIFAR Inc10}}& 
        \multicolumn{2}{c}{\textbf{TinyImg Inc20}}&
        \multicolumn{2}{c}{\textbf{ImgR Inc20}} \\
         &  Last. & Avg. &  Last. & Avg. &  Last. & Avg.\\
        \midrule
        Dual \textit{w/ Hop-\textcolor{gray}{odd}} \dag   & 78.76&\textbf{86.14} &52.04&60.47& 64.90 & 71.62 \\
        \textbf{MoZO} \dag & \textbf{78.84}& 86.11&\textbf{52.12}&\textbf{60.67}& \textbf{64.92} & \textbf{71.67}  \\
        \midrule
       Dual \textit{w/ Hop-\textcolor{gray}{even}} \dag  & 79.22& \textbf{86.27}&52.66&61.32& 65.68 & 71.67 \\
        \textbf{MoZO} \dag  &\textbf{79.30}& 86.23&\textbf{53.54}&\textbf{61.99}& \textbf{65.89} & \textbf{71.98}  \\
        \bottomrule
    \end{tabular}
   }
   \caption{Effect of MoZO optimization on performance under LoRA setting (represented by \dag). Dual \textit{w/ Hop-\textcolor{gray}{odd}/\textcolor{gray}{even}} indicates the results of adopting \textit{Hop-\textcolor{gray}{odd}} and \textit{Hop-\textcolor{gray}{even}} layer-wise ZO in dual branch.}
   \label{lorazostratgety}
\end{table}

\section{Exploring the effect MoZO Optimization under LoRA Setting}
\label{secb}

In the main paper, we explore the discrepant behavior of ZO across different modality branches, which motivates our proposed improvements for applying ZO in the VLCL setting. We replace the baseline architecture with LoRA and conduct the same investigation under identical experimental settings to revisit this phenomenon. Table~\ref{lorazostratgety} demonstrates that applying gradient regularization and reducing the perturbation in the vision branch leads to further improvements in most experimental results, this indicates that our MoZO method remains effective under the LoRA setting. We further analyze this method by comparing the gradient variance in \textit{Hop-\textcolor{gray}{odd}} vs. \textit{Hop-\textcolor{gray}{even}} on CIFAR dataset shown in Figure~\ref{fig:lorasuppresszo}, it can be observed that the proposed method has a positive impact, as the gradient variance becomes more stable. These results are broadly consistent with the conclusions presented in the main paper, further demonstrating the effectiveness of the proposed method.

\begin{figure}[hbpt]
    \centering
    \small
    \begin{minipage}{\columnwidth}
        
        \begin{subfigure}[b]{0.48\columnwidth}
            \centering
            \includegraphics[width=\textwidth]{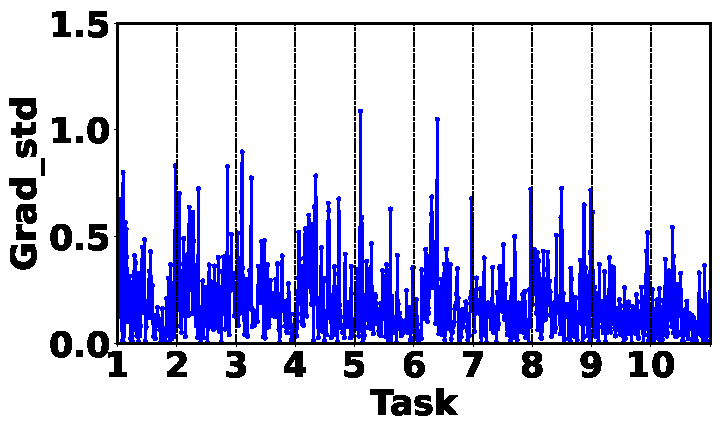}
            \caption{\textit{Hop-\textcolor{gray}{odd}} \dag}
            \label{fig:loraodd}
        \end{subfigure}
        \begin{subfigure}[b]{0.48\columnwidth}
            \centering
            \includegraphics[width=\textwidth]{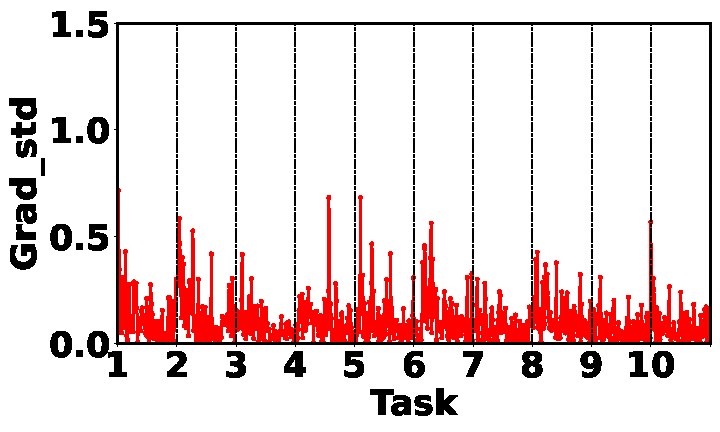}
            \caption{\textit{Hop-\textcolor{gray}{odd}} \dag}
            \label{fig:comloraodd}
        \end{subfigure}
        \\
        \begin{subfigure}[b]{0.48\columnwidth}
            \centering
            \includegraphics[width=\textwidth]{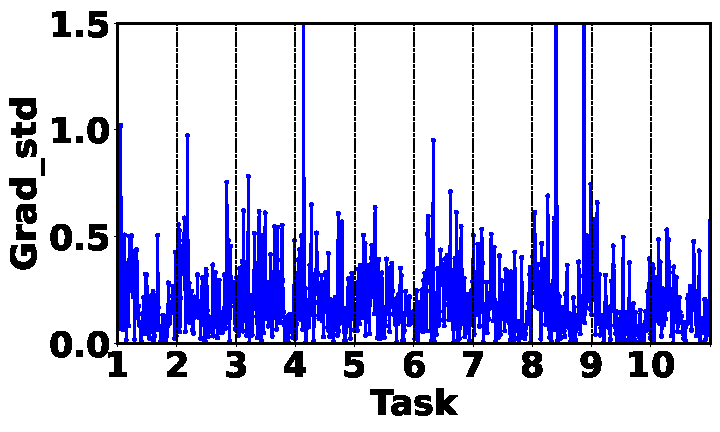}
            \caption{\textit{Hop-\textcolor{gray}{even}} \dag}
            \label{fig:loraeven}
        \end{subfigure}
        \begin{subfigure}[b]{0.48\columnwidth}
            \centering
            \includegraphics[width=\textwidth]{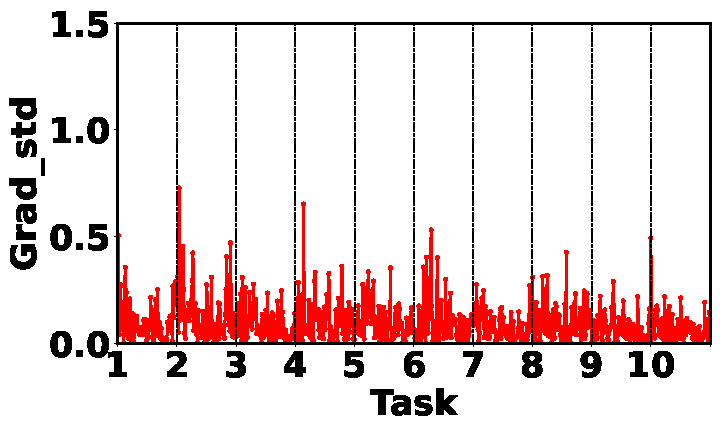}
            \caption{\textit{Hop-\textcolor{gray}{even}} \dag}
            \label{fig:comloraeven}
        \end{subfigure}
    \end{minipage}
    \caption{Analyzing the effect of MoZO optimization in \textit{Hop-\textcolor{gray}{odd}} vs. \textit{Hop-\textcolor{gray}{even}} for LoRA setting (represented by \dag). Blue shows original results, while red shows the positive impact attributed to vision discrepancy.}
    \label{fig:lorasuppresszo}
\end{figure}

\end{document}